\documentclass{article}

\newif\ifpreprint
\preprinttrue

\usepackage[preprint, nonatbib]{neurips_2022}

\usepackage[utf8]{inputenc} 
\usepackage[T1]{fontenc}    
\usepackage{hyperref}       
\usepackage{url}            
\usepackage{booktabs}       
\usepackage{amsfonts}       
\usepackage{nicefrac}       
\usepackage{microtype}      
\usepackage{xcolor}         

\usepackage{graphicx}
\usepackage{amsmath}
\usepackage{xspace}
\usepackage{footnote}
\makesavenoteenv{tabular}
\makesavenoteenv{table}

\newcommand{\gbs}[0]{\ensuremath{\vec{G}_{B_{\text{small}}}}\xspace}
\newcommand{\gbl}[0]{\ensuremath{\vec{G}_{B_{\text{big}}}}\xspace}

\newcommand{\bbs}[0]{\ensuremath{{B_{\text{small}}}}\xspace}
\newcommand{\bbb}[0]{\ensuremath{{B_{\text{big}}}}\xspace}

\newcommand{\bs}[0]{\ensuremath{\mathcal{B}_\text{simple}}\xspace}
\newcommand{\bn}[0]{\ensuremath{\mathcal{B}_\text{noise}}\xspace}

\newcommand{\vt}[0]{\ensuremath{V_\text{targ}}\xspace}

\newcommand{\lp}[0]{\ensuremath{\lambda_\pi}}
\newcommand{\lv}[0]{\ensuremath{\lambda_V}}

\usepackage{algorithmicx}
\usepackage[noend]{algpseudocode}
\usepackage{algorithm}

\usepackage{xcolor}
\hypersetup{
    colorlinks,
    linkcolor={red!50!black},
    citecolor={blue!50!black},
    urlcolor={blue!80!black}
}

\title{DNA: Proximal Policy Optimization with a Dual Network Architecture}

\author{%
  Matthew Aitchison\\
  The Australian National University\\
  \texttt{matthew.aitchison@anu.edu.au} \\
  \And
  Penny Sweetser\\
  The Australian National University\\
}

\begin{document}

\maketitle




\begin{abstract}
This paper explores the problem of simultaneously learning a value function and policy in deep actor-critic reinforcement learning models. We find that the common practice of learning these functions jointly is sub-optimal due to an order-of-magnitude difference in noise levels between the two tasks. Instead, we show that learning these tasks independently, but with a constrained distillation phase, significantly improves performance. Furthermore, we find that policy gradient noise levels decrease when using a lower \textit{variance} return estimate. Whereas, value learning noise level decreases with a lower \textit{bias} estimate. Together these insights inform an extension to Proximal Policy Optimization we call \textit{Dual Network Architecture} (DNA), which significantly outperforms its predecessor. DNA also exceeds the performance of the popular Rainbow DQN algorithm on four of the five environments tested, even under more difficult stochastic control settings.
\end{abstract}

\section{Introduction}

Combining deep neural networks with reinforcement learning has produced impressive results on challenging problems, such as playing Chess \cite{silver2018general}, Atari games \cite{mnih2015human}, and robotics tasks \cite{schulman2017proximal}. However, results have bifurcated between two competing approaches: Q-learning-based approaches that require learning an (action conditioned) value estimate and actor-critic policy gradient (AC-PG, or just PG)\footnote{For simplicity, we refer to actor-critic policy gradient as policy gradient, even though some early policy gradient approaches, such as REINFORCE \cite{williams1992simple}, do not learn a value function.} approaches that learn both a policy and value estimate. PG offers many theoretical advantages over Q-learning, such as natural support for continuous control problems and the ability to learn stochastic policies. Until recently, PG approaches \cite{mnih2016asynchronous, schulman2017proximal, cobbe2021phasic}, while strong on continuous control problems, have under-performed Q-learning approaches on vision-based tasks \cite{mnih2015human, hessel2018rainbow, badia2020agent57}, restricting the use of PG in this domain.

This paper aims to close the gap between PG and Q-learning methods by showing that the common practice of jointly learning value and policy with shared features negatively affects the algorithm's performance.\footnote{PPO \cite{schulman2017proximal}, A3C \cite{mnih2016asynchronous}, and Muslie \cite{hessel2021muesli} all use a joint network when training on vision-based discrete action environments. PPG \cite{cobbe2021phasic} uses a dual network with an auxiliary task, which we discuss in more detail in Section \ref{sec:background}} We also demonstrate an order-of-magnitude difference in noise levels between these two tasks and argue that this makes these two tasks poorly aligned.

In light of this result, we introduce a new algorithm, based on Proximal Policy Optimization (PPO) \cite{schulman2017proximal}, called Dual Network Architecture (DNA). We test DNA empirically on a subset of the Arcade Learning Environment (ALE) \cite{bellemare2013arcade}, Atari-5 \cite{aitchison2022atari}.  Our results show a strong increase in performance on this benchmark compared to both PPO and another dual network model Phasic Policy Gradient (PPG) \cite{cobbe2021phasic}. Our model also outperforms Rainbow DQN \cite{hessel2018rainbow} on four of the five games tested, even under more challenging environmental settings. 

We summarize our contributions. First, we provide empirical results showing an order-of-magnitude difference in the noise scale between the policy gradient and the value loss gradient. Second, we give evidence for the benefit of using low \textit{bias} value estimates for value learning, but low \textit{variance} estimates for advantage estimates. Finally, we introduce and justify our dual network constrained distillation algorithm DNA with empirical results on ALE.
 
\section{Preliminaries and Related Work}
\label{sec:background}

\paragraph{Proximal Policy Optimization} Our algorithm builds on the well-established Proximal Policy Optimization (PPO) algorithm \cite{schulman2017proximal}. PPO is a policy gradient, reinforcement learning algorithm, with many of the advantages of its predecessor Trust Region Policy Optimization \cite{schulman2015trust}, while being much simpler to implement. Various aspects of PPO have been studied in recent years, such as the impact of implementation choices \cite{engstrom2019implementation, shengyi2022the37implementation}, the ability to generalize \cite{cobbe2021phasic, raileanu2021decoupling}, and performance under multi-agent settings \cite{yu2021surprising}. There has also been a growing understanding that the value and policy learning tasks involved in actor-critic models like PPO have important asymmetries \cite{raileanu2021decoupling, cobbe2021phasic}. Our work builds on this by investigating how differences in value and policy noise levels can be accommodated to improve PPO's performance on video-based discrete action tasks.


\paragraph{TD($\lambda$) Return Estimation}


Policy gradient algorithms often make use of a value estimate as a baseline \cite{mnih2016asynchronous}, as well as for estimating the value of truncated trajectories \cite{cichosz1994truncating}. To better facilitate control over the noise levels for policy and value learning, our work makes use of two different return estimations, both using TD($\lambda$) \cite{sutton1988learning}. Given value estimates $V(\cdot)$ from the value network we define the n-step value estimate for some state $s_t$ taken at time $t$, and their exponentially weighted sum as

\begin{align}
    \text{NSTEP}^{(\gamma, k)}(s_t) &:= \sum_{i=0}^{k-1} \gamma^i r_{t+i} + \gamma^k V(s_{t+k}),\\
    \text{TD}^{(\gamma, \lambda)}(s_t) &:= (1-\lambda) \sum_{k=1}^{\infty} \lambda^{k-1} \text{NSTEP}^{(\gamma, k)}(s_t).
\end{align}

For values of $\lambda$ close to $1$, more weight is assigned to longer n-step return estimates, and less to shorter ones. There has been a long-standing belief that shorter n-step returns generate more biased estimates, whereas longer n-step estimates, due to summing over many stochastic rewards, have higher variance.\footnote{While there are cases (i.e. where rewards are negatively correlated) where this is not the case, empirical experiments confirm the intuition that longer n-step estimates generally have higher variance and lower bias.} For a more thorough discussion on this topic see \cite{kearns2000bias}.

\paragraph{Noise Scale.} Noise scale is a measure of noise-to-signal in stochastic gradient descent (SGD). While SGD provides unbiased gradient estimates, these estimates are typically very noisy. Each sampled gradient estimate $\hat{G}$ can be thought of as a sum of the true gradient $\vec{G}$ and some noise vector $\vec{\sigma}$. A method for estimating the ratio of the magnitude of this implied noise vector to the magnitude of the true gradient was proposed by \cite{mccandlish2018empirical} who show that their efficient-to-calculate \textit{simple noise scale} is a good match for the noise scale. They also show that the noise scale provides useful information about the choice of mini-batch size to use when estimating gradients.

\paragraph{Phasic Policy Gradient.} Most similar to our work is Phasic Policy Gradient (PPG) \cite{cobbe2021phasic}. Like DNA, PPG has three distinct phases during training and uses two independent networks. However, there are several important distinctions in our work. First, we forgo the large replay buffer, relying instead on learning entirely from recent experience. This significantly reduces the memory requirements of our algorithm. Second, we make use of a distillation phase rather than an auxiliary phase.\footnote{PPG refer to their third phase as a distillation phase, however, because the update trains both the policy and value networks on value estimates generated from rollouts, it is better thought of as an auxiliary task. We discuss this in more detail in Section \ref{sec:distillation}} Finally, our work reduces gradient noise by using two different return estimators calibrated for the characteristics of each task. We assess the impact of these differences on the performance of the agent in Section \ref{sec:results}.

\section{The Noise Properties of Value Learning and Policy Gradient}
\label{sec:noise}

Here we examine the noise properties of the policy gradient and value learning tasks and show that the policy gradient has a much higher noise level than value learning. This implies different challenges in the optimization problem. Specifically, that policy should be learned with a much larger mini-batch size than value learning and that return estimates should be adapted appropriately.

\subsection{Motivating Example}

\begin{figure}[h]
    \centering
    \includegraphics[height=4.3cm]{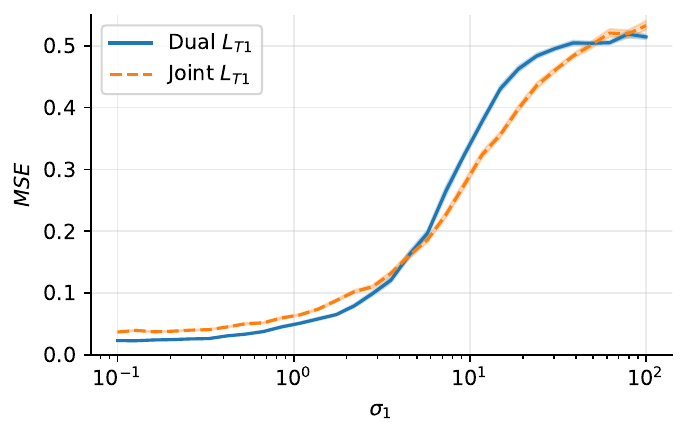}
    \includegraphics[height=4.3cm]{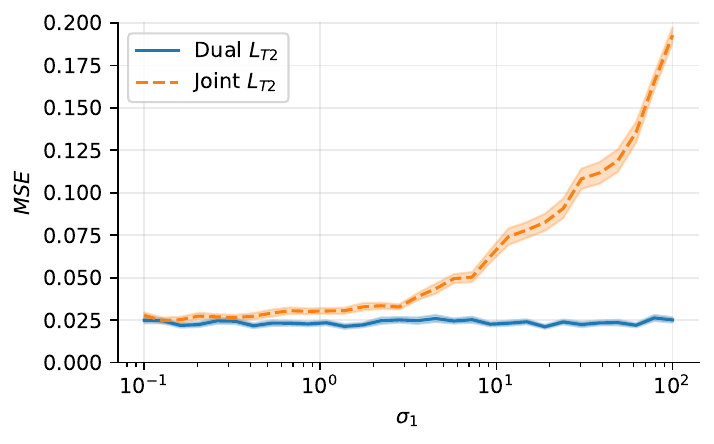}
    \caption{A depiction of the problem of destructive interference on a toy problem. $L_{T1}$ and $L_{T2}$ refer to the mean-squared-error on the first and second task respectively. Shading indicates standard error over 100 seeds.}
    \label{fig:toy_noise}
\end{figure}

To motivate our investigation into noise levels, we consider the task of learning two independent functions $F_1(x) := \sin(5x)+\text{N}(0, \sigma_1^2)$ and $F_2(x) := \cos(5x)+\text{N}(0, \sigma_2^2)$, where $\text{N}(\mu,\sigma^2)$ is Gaussian noise with mean $\mu$ and standard deviation $\sigma$,  over the domain $[-\pi, \pi]$. We fix $\sigma_2$ = 1, and vary $\sigma_1$ on a log scale from $0.1$ to $100$. We trained two 3-layer multi-layer perceptions (MLPs) on this problem.\footnote{It is worth noting that we did not find the same result when using shallow networks. That is, dual and joint models performed similarly when using a 2-layer MLP.}  The first was a joint model, using a first hidden layer of 1024 units and a second hidden layer of 2048 units, followed by two output heads. Our second model was a dual network consisting of two independent MLPs with 1024 units on each hidden layer.\footnote{This number of hidden units was chosen so that both models had the same number of parameters.} All models used ReLU activations in between linear layers, and error was measured as mean-squared-error (MSE) between the predicted value and the \textit{noise free} true value. The results are presented in Figure \ref{fig:toy_noise}. At low noise levels, the tasks did not interfere, but as the noise of the first task increased, performance on the second task eventually degraded for the joint network, but not the dual network. While simple, this experiment demonstrates the impact one noisy task can have on another when learned jointly by a deep neural network.

\subsection{Noise Scale in Reinforcement Learning}


To assess the differences in noise levels in our experiments, we measure the \textit{gradient noise scale} of the policy gradient and value loss gradient. We did this by calculating an estimate of the noise scale, developed by \cite{mccandlish2018empirical} called \textit{simple noise scale} defined as
\begin{align}
    \bs := \frac{\text{tr}(\Sigma)}{|\vec{G}|^2},
\end{align}
where $\Sigma$ is the gradient covariant matrix, and $\vec{G}$ is the true gradient. For convenience we also use the notation $\sigma := \sqrt{\bs}$, which can interpreted as the ratio of the length of the implied noise vector to the length of the true gradient vector. That is, a kind of noise-to-signal ratio. As \cite{mccandlish2018empirical} have shown, a low bias  estimate of \bs can be found efficiently by generating gradient estimates \gbs using a small mini-batch of size \bbs as well as the gradient \gbl using a large mini-batch of size \bbb, then calculating

\begin{align}
     |\mathcal{G}|^2 &:= \frac{1}{\bbb - \bbs}\large(  \bbb |\gbl|^2 - \bbs |\gbs|^2   \large) \\
     \mathcal{S} &:= \frac{1}{1/\bbs - 1/ \bbb} \large( |\gbs|^2 - |\gbl|^2   \large),
\end{align}

where $\mathbb{E}[|\mathcal{G}|^2] = |\vec{G}|^2$ and $\mathbb{E}[\mathcal{S}] = \text{tr}(\Sigma)$. 

To evaluate the noise scale of value learning and policy gradient we trained our dual network model (described fully in section \ref{sec:dna}) on the three Atari games from the Atari-3 validation set \cite{aitchison2022atari}. Like \cite{mccandlish2018empirical}, we found it necessary to smooth out noise by maintaining an exponential moving average of $|\gbl|^2$. We used $\bbs=32$ and $\bbb=16,384$ in all our experiments. Pseudocode for this procedure is given in Appendix \ref{app:noise_estimate}.


\begin{figure}[t]
    \centering
    \includegraphics[width=1.0\textwidth]{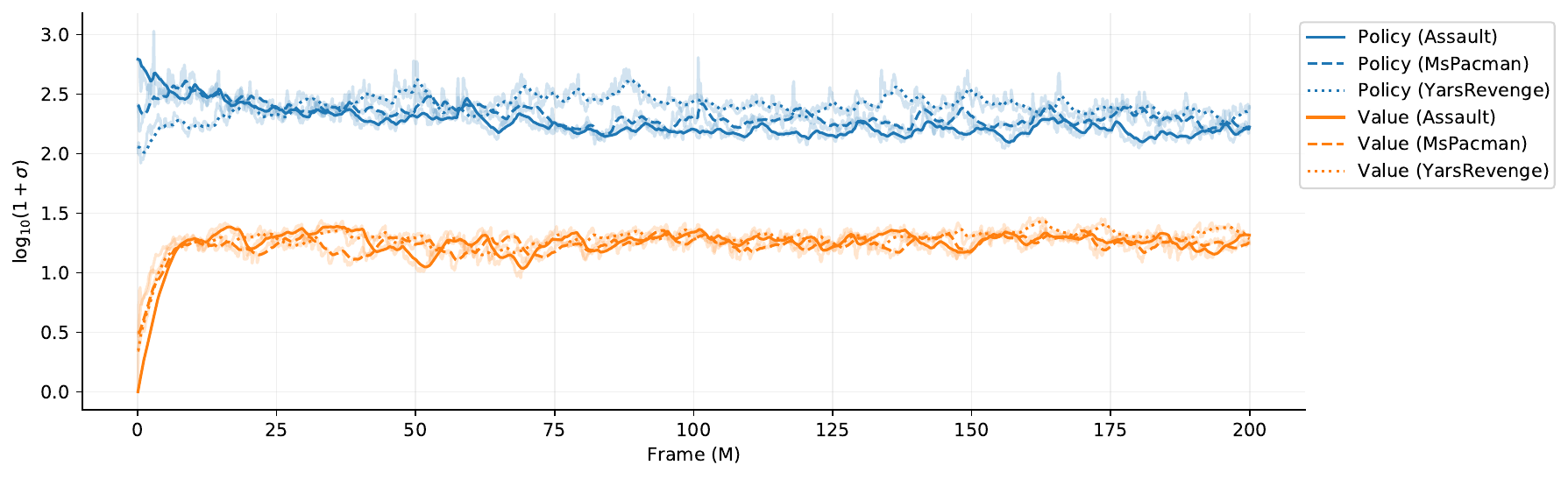}
    \caption{Noise scale for the two learning objectives. For clarity all values are further smoothed using an exponential rolling average (with non-smoothed values presented in a lighter shade). Smoothing may introduce bias, as while $|\mathcal{G}|^2$ and $\mathcal{S}$ are unbiased estimators, their ratio may not be.}
    \label{fig:noise_scale}
\end{figure}

The results are presented in Figure \ref{fig:noise_scale}. We found that policy noise ($\sigma_\pi$) reduced by about half during training and that value noise ($\sigma_v$) remained constant. Both noise levels varied very little between the three environments. A large variation was observed between the value loss and policy gradient, consistent between environments. The noise levels, measured at the end of training and averaged over all three environments, were found to be $\sigma_\pi=220.5$ and $\sigma_V=17.5$, representing a $12.6$ times difference.

\section{Dual Network Architecture}
\label{sec:dna}

Based on our experimental results in Section \ref{sec:noise}, we propose an architecture which takes into account the large difference in noise levels between the two tasks. This architecture consists of three improvements to PPO. First, to reduce negative interference from the noisy policy gradient, policy and value should be learned by independent networks with different hyperparameters (e.g. training epochs and mini-batch size). Second, the variance/bias trade-off in the return estimations should be calibrated to each task. Finally, a constrained distillation phase should be applied to take advantage of any constructive interference between the two tasks.\footnote{The source code used to generate the results in this paper is provided in the supplementary material. We also provide an implementation of our algorithm at \url{https://github.com/maitchison/PPO/tree/DNA}.}

\subsection{Independent Networks}



\begin{figure}[t]
    \centering
    \includegraphics[width=0.6\textwidth]{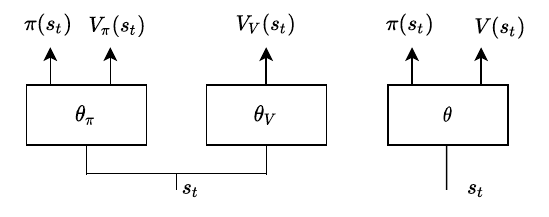}
    \caption{Architecture of DNA (left) compared to a single network setup (right).}
    \label{fig:dna_arch}
\end{figure}

Important to DNA is the use of a dual network architecture.\footnote{By dual network, we mean two independent networks. Not to be confused with a dual-head network (which we refer to as a joint network), or with dualing networks \cite{wang2016dueling}.} Not only does this setup allow for learning of the value and policy without destructive interference, it also enables a specialized set of hyperparameters to be calibrated for the distinct tasks of policy and value learning. %
Specifically, the calibration of the mini-batch size, which it has been suggested is best set roughly proportional to $\sigma$ \cite{mccandlish2018empirical}. The policy network outputs a policy $\pi$ and a value estimate $V_\pi$, whereas the value network only outputs a value $V_V$, as depicted in Figure \ref{fig:dna_arch}.


\subsection{Decoupled Return Estimation}

To better facilitate control over the noise levels for policy and value learning, DNA makes use of two different return estimations, both using TD($\lambda$) \cite{sutton1988learning} as follows
\begin{align}
    V_\text{targ}(s_t) &:= \text{TD}^{(\gamma, \lambda_V)}(s_t), \\
    V_\text{adv}(s_t) &:= \text{TD}^{(\gamma, \lambda_\pi)}(s_t),
\end{align}

where $\lambda_V$ and $\lambda_\pi$ are hyperparameters controlling the variance / bias trade-off of each estimate. The $V_\text{targ}$ estimates are used as targets for training the value function, whereas $V_\text{adv}$ are used for advantage estimates used for policy gradient estimates given by


\begin{align}
\hat{A}_t &:= V_\text{adv}(s_t) - V(s_t).
\end{align}

This formulation of the advantages is equivalent to the general advantage estimation (GAE) \cite{schulman2015high}, that is, for any $\lambda \in [0..1], \gamma \in [0..1)$  we have,\footnote{A proof for this claim is provided in Appendix \ref{app:td}}
\begin{align}
    \hat{A}_t^{\text{GAE}(\gamma, \lambda)} &= \text{TD}^{(\gamma, \lambda)}(s_t) - V(s_t).
\end{align}
Because of this, it is common when using PPO to generate value targets, $V_\text{targ}$, for the value function by adding the network's value estimates to the (unnormalized) advantages, implicitly settings $\lv=\lp$.\footnote{For example, see line 65 of the popular baselines implementation of PPO. \url{https://github.com/openai/baselines/blob/ea25b9e8b234e6ee1bca43083f8f3cf974143998/baselines/ppo2/runner.py} } We hypothesize that due to bootstrapping, estimates used for \vt would benefit from being \textit{low bias}, conversely, because estimates used for advantage estimation contribute to the noise of the (already very noisy) policy gradient, they would benefit from being \textit{low variance}. That is, the bootstrapping process may cause symmetrically distributed error, but not biased error, to cancel out. This suggests setting $\lp < \lv$, a thought which we test empirically in section \ref{sec:results_re}.

\subsection{Distillation}
\label{sec:distillation}

Because value learning is a simpler task in terms of noise level, it makes sense to try to transfer some of the knowledge learned by the value network to the policy network. If the value network learns to identify important features in the environment, we would like the policy network to identify those features also. To facilitate this, we employ a constrained distillation update \cite{hinton2015distilling}. Distillation between two identical networks has the property that the student's target function (since an identical network generated it) is guaranteed to be in the model's solution space. Also, because the distillation targets are deterministic,\footnote{That is to say, $V_\text{targ}(s_t)$ are samples drawn from the random variable $R(s_t)$, where $R$ is a (noisy) return estimator, whereas $V(s_t)$ is a deterministic function that produces an estimate of $\mathbb{E}[R(s_t)]$.} the process is also very low noise, which we verify in Appendix \ref{app:distil_noise}.

We contrast this with the auxiliary update of PPG which trains both the value function and the policy on an auxiliary task. This auxiliary task equates to learning $\frac{1}{N}\sum_{i=0}^{N-1}{V_\text{targ}^{\pi_{-i}}}$ where $V_\text{targ}^{\pi_{-i}}$ is a value estimate for the policy from $i$ updates prior. For quickly changing policies, and large $N$, these updates may cause tension with the value estimates of the current policy.\footnote{A better way to do auxiliary tasks would be to train a separate value network output head, dedicated to the auxiliary task. This way the agent could learn the value of the current policy, along with the moving average value. We believe, however, that the better solution to overfitting the value function to recent data is simply to train less. An idea which we discuss further in Section \ref{sec:discussion}.}

We use $V_V(s)$ as the distillation targets, with the input states $s$ being taken from the current rollout. However, unlike policy and value training, distillation state inputs need not be on-policy. We explored other distillation targets, covered in Appendix \ref{app:distil_targets}, but found $V_V(s)$ to be the best of those tried. Distillation is performed, like PPG's auxiliary task, using mean squared error, and under a soft constraint on the policy network's policy, specifically
\begin{equation}
    L_t^{D}(\theta) := \mathbb{\hat{E}}_t \big[(V_\pi(s_t)-V_V(s_t))^2 \big] +
    \beta \cdot \mathbb{\hat{E}}_t\large[ \text{KL}(\pi_\text{old}(\cdot|s_t), \pi(\cdot|s_t)) \big]
\end{equation}
where $\beta$ is the policy constraint coefficient, and $\pi_\text{old}$ is a copy of the policy before the distillation update. During updates gradients are only propagated through the policy network, and not the value network.

\subsection{Training}

Like Phasic Policy Gradient (PPG) \cite{cobbe2021phasic}, DNA splits training into three distinct phases (policy, value, and distil), but unlike PPG, rather than using a large replay buffer, we perform all updates on-policy on the current batch of rollout data. Each of the three phases optimizes a single objective, for some number of epochs, using its own optimizer, with a unique set of hyperparameters. The optimization objective for the policy phase of DNA is the clipped surrogate object from PPO \cite{schulman2017proximal} including the entropy bonus 
\begin{equation}
    L^{\text{CLIP}}_t := \hat{\mathbb{E}}_t \bigg[ \text{min}( \rho_t(\theta)\hat{A_t}, \text{clip}(\rho_t(\theta), 1-\epsilon, 1+\epsilon)\hat{A}_t) + c_\text{eb} \cdot \text{S}[\pi(s_t)] \bigg]
\end{equation}
where $\text{S}$ is the entropy in nats, $\rho_t$ is the ratio $\frac{\pi(a_t|s_t)}{\pi_\text{old}(a_t|s_t)}$ at time $t$, $\epsilon$ is the clipping coefficient, and $c_\text{eb}$ is the entropy bonus coefficient. For the value phase, we we use the squared-error value loss, 
\begin{equation}
    L_t^{VF} := \mathbb{\hat{E}}_t \big[ (V_V(s_t)-V_{\text{targ}}(s_t))^{2} \big].
\end{equation}
In summary, the DNA algorithm separates the tasks of policy learning and value learning into a network to handle the high noise of policy learning and a separate network to handle the lower noise task of value learning. Knowledge from the value learning network is transferred to the policy network through a separate constrained distillation phase, which allows for constructive interference between the learning tasks while minimizing the destructive. We formalize the algorithm as follows.

\begin{algorithm}[H]
    \caption{Proximal Policy Optimization with Dual Network Architecture}
    \label{alg:ppo_dna}
    \begin{algorithmic}[1]
    \Procedure{PPO-DNA}{}
    \State \textbf{Input} $N \in \mathbb{Z^+}$ rollout horizon
    \State \textbf{Input} $A \in \mathbb{Z^+}$ number of agents
    \State \textbf{Input} $\pi$ the initial policy.
    \For{$i = 1$ to ...}
        \For{$a = 1$ to $A$}
            \State{Run policy $\pi$ in environment $a$ for $N$ timesteps}
        \EndFor
        
        \State{Compute $V_\text{targ} \gets \text{TD}^{(\gamma, \lambda_V)}$}
        
        \State{Compute $V_\text{adv} \gets \text{TD}^{(\gamma, \lambda_\pi)}$}
        
        \State{Compute $\hat{A} \gets \text{TD}^{(\gamma, \lambda_\pi)} - V_\text{adv}$}
    
        \For{$j = 1$ to $E_\pi$}
            \State{Optimize $L^{CLIP}$ wrt $\theta_{\pi}$}
        \EndFor
    
        \For{$j = 1$ to $E_V$}
            \State{Optimize $L^{VF}$ wrt $\theta_V$}
        \EndFor
        
        \State{$\pi_\text{old} \gets \pi $ }
    
        \For{$j = 1$ to $E_D$}
            \State{Optimize $L^{D}$ wrt $\theta_\pi$}
        \EndFor
    \EndFor
    \State \textbf{Output} $\pi$
    \EndProcedure
    \end{algorithmic}
\end{algorithm}

\section{Evaluation}
\label{sec:evaluation}



To evaluate our algorithm's performance, we used the Atari-5 benchmark \cite{aitchison2022atari}. Scores in Atari-5 are generated using a weighted geometric average over five specific games and produce results that correlate well with the median score if all 57-games had been evaluated. This allowed us to perform multiple seeded runs and defined a clear training and test split between the games. In all cases, we fit hyperparameters to the 3-game validation set and only used the 5-game test set for final evaluations. 

We opted for the more difficult stochastic ALE settings recommended as best practice by \cite{machado2018revisiting}. However, to better understand our results in the context of prior work, we also provide results under the simpler deterministic settings in Appendix \ref{app:easy}, and additionally provide for reference a single evaluation on the full 57-game set in Appendix \ref{app:full}. Unless otherwise specified, agents were scored according to their average performance over the previously completed 100-episodes at the end of training.

A coarse hyperparameter sweep found initial hyperparameters for our model on the Atari-3 validation set. Notably, we found the optimal mini-batch size for value and distillation to be the minimum tested (256), while the optimal mini-batch size for policy was the largest tested (2048). For optimization, we used Adam \cite{kingma2014adam}, over the standard 200 million frames.\footnote{That is 50 million environment interactions.} Full hyperparameter details for our experiments are given in Appendix \ref{app:hyperparameters}.

In order to understand the impact of the hyperparameters introduced by our algorithm, namely $E_\pi, E_V, E_D$ and $\lambda_\pi, \lambda_V$, we performed the following experiments on the Atari-3 validation set. We started by setting $E_\pi = E_V = E_D = 2$, then searched over $E_V \in [1,2,3,4]$, and selected the best $E_V$. We then searched over $E_\pi$ and $E_D$ while keeping the other two parameters constant. We also performed a similar experiment on the impact of $\lambda_V$ and $\lambda_\pi$ by setting $\lambda_V$ to 0.95 then sweeping across $\lambda_\pi \in [0.6, 0.8, 0.9, 0.95, 0.975]$, then setting $\lambda_V$ to the best $\lambda_\pi$ and repeating. This process allowed us to verify if the best settings occur at $\lambda_V=\lambda_\pi$ or at $\lambda_V \neq \lambda_\pi$. During our $\lambda$ tuning, we also recorded noise levels for one of the seeds, the results of which we discuss in Section \ref{sec:results_re}.

We evaluated our algorithm DNA, against PPO, and PPG. All models used the `NatureCNN' encoder from \cite{mnih2015human}. However, because DNA and PPG use two networks and thus twice the parameters, we doubled the number of channels of the PPO encoder which we refer to as PPO (2x) and which is very similar to the encoder used by \cite{badia2020agent57}. All models have have approximately 3.5 million parameters in total. For fairness we repeated the same sweep on PPO for the number of training epochs ($E_\text{PPO}$) and $\lambda$ used for GAE ($\lambda_\text{PPO}$). To verify that our settings did not inadvertently degrade the performance of PPO we also tested against the settings used by \cite{schulman2017proximal} and refer to this as PPO (original). These results are included in the ablation study in Appendix \ref{app:ablation}.

\section{Main Experimental Results}
\label{sec:results}

In this section we present the results of our main study comparing DNA to PPO and PPG on Atari-5. We also include supplementary experiments on MuJoCo \cite{todorov2012mujoco} and Procgen \cite{cobbe2020leveraging} in Appendix \ref{app:mujoco}, and \ref{app:procgen} respectively. 

\subsection{Impact of Training Epochs}

\begin{figure}[t]
    \centering
    \includegraphics[width=0.32\textwidth]{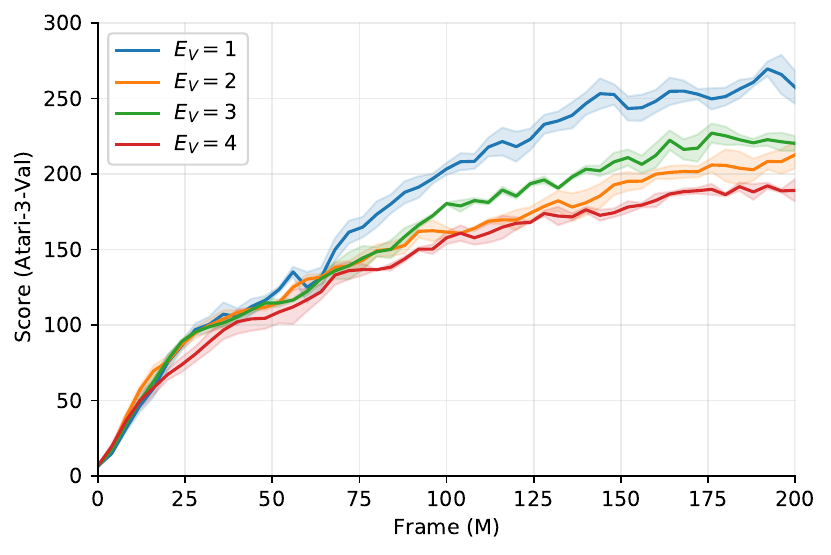}
    \includegraphics[width=0.32\textwidth]{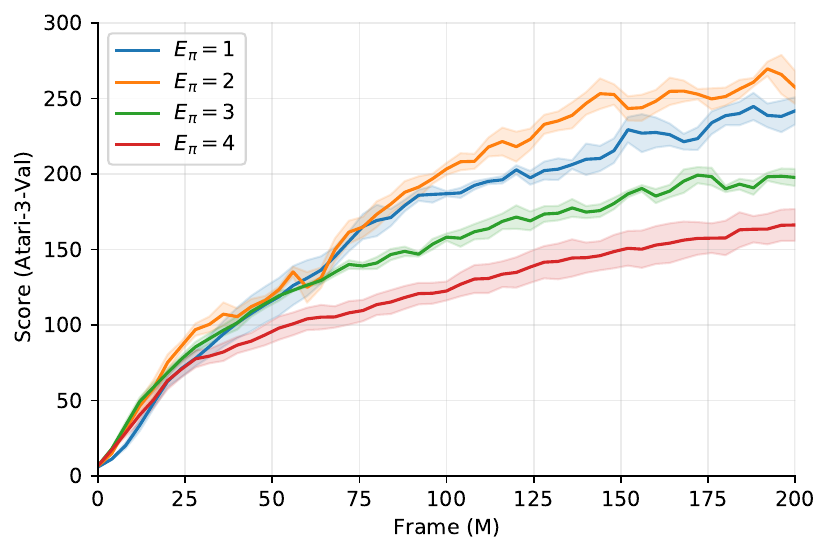}
    \includegraphics[width=0.32\textwidth]{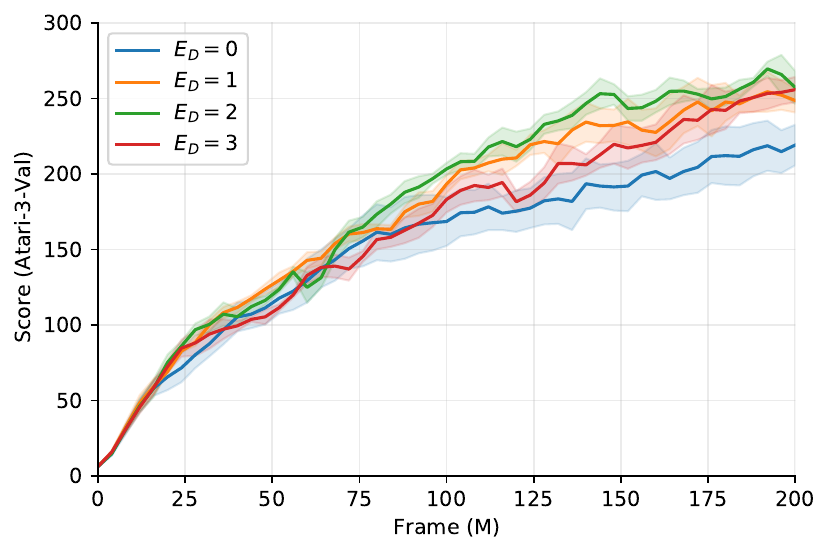}
    \caption{Training curves for DNA with various epochs. The DNA algorithm is most sensitive to the number of policy epochs, and is optimal on our validation set with $E_\pi=2$. Applying policy updates 3 or 4 times significantly reduces the agents performance.}
    \label{fig:dna_tuning}
\end{figure}

We start by investigating the impact of the $E_\pi$, $E_V$, and $E_D$ hyperparameters. We found that using \textit{less} than three policy epochs for DNA dramatically increased the performance of the agent while also decreasing the computation required to train the agent (Figure \ref{fig:dna_tuning}). We also found that DNA was robust to the choice of value and distillation epochs but that overtraining on value ($E_V=4$) marginally decreased performance. This was not the case for distillation updates. The distillation phase demonstrated some benefit as indicated by the $E_D=0$ results under-perform the others. We selected $E_\pi=2, E_V=1, E_D=2$ as the optimal settings and used these for the remainder of the experiments. These settings require a total of four updates for the policy network, but only one update for the value network (we discuss this further in Section \ref{sec:update_ratio}). The training curves also suggest DNA may benefit from training beyond the standard 200 million frames. 

\subsection{Return Estimation}
\label{sec:results_re}

\begin{figure}[!t]
    \centering
    \includegraphics[height=1.3in]{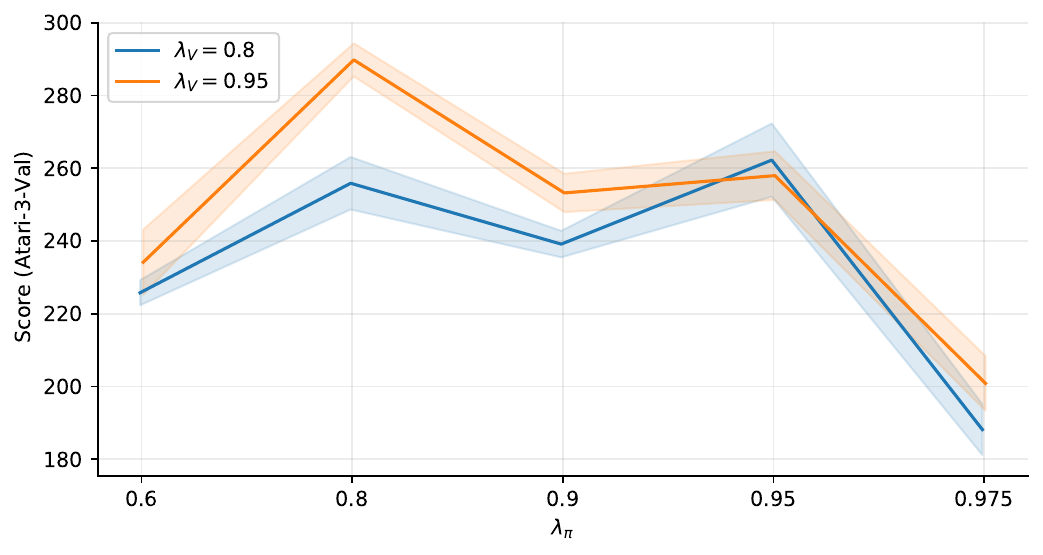}
    \includegraphics[height=1.3in]{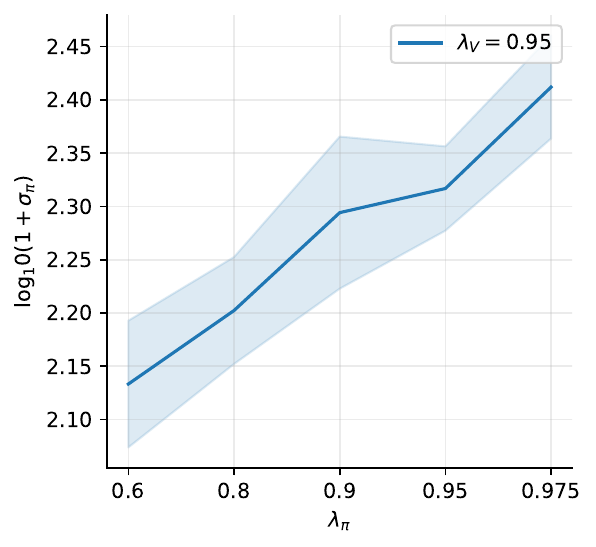}
    \includegraphics[height=1.3in]{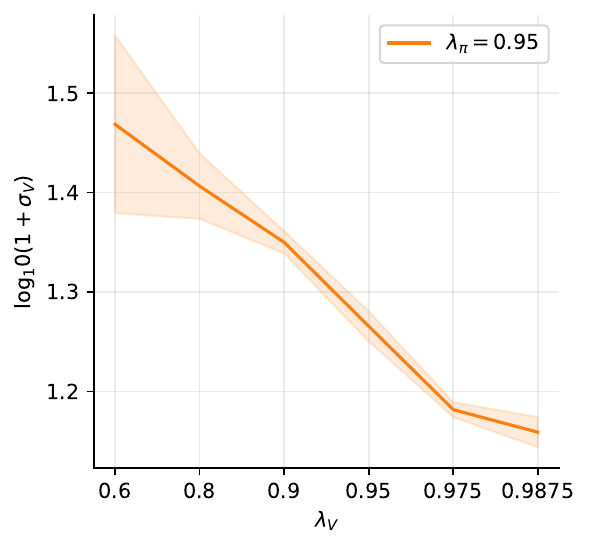}
    \caption{Performance of DNA across a range of $\lambda$ values for return estimations. Left: The optimal settings for $\lambda_V$ and $\lambda_\pi$ differed. Shading indicates standard error over 5 seeds. Mid/right: Noise levels from a single seed, $\sigma_\pi$ increased with $\lambda_\pi$ where as $\sigma_V$ decreased with $\lambda_V$. Shading indicates standard deviation over the three environments.}
    \label{fig:dna_lambda}
\end{figure}

We found that the choice of $\lambda_\pi$ made a large difference to the performance of our agent. As was hypothesised the optimal settings for $\lambda_\pi$ differed to that of $\lambda_V$ with $\lambda_\pi$ preferring a lower variance, higher bias value of $0.8$, while the value targets, $\lambda_V$ preferred a lower bias, higher variance setting of $0.95$ (Figure \ref{fig:dna_lambda}). Notable is that the non-homogeneous setting $\lambda_V=0.95, \lambda_\pi=0.8$ outperformed both homogeneous choices $\lambda_\pi=\lambda_V=0.8$ and $\lambda_\pi=\lambda_V=0.95$. We found, as expected, that policy gradient noise can be reduced by selecting a low value for $\lambda_\pi$. However, against our expectations, setting $\lambda_V$ lower actually increased, not decreased the noise level of the value learning task. Once we discovered this we ran addition experiments for a broader choice of $\lambda_V$, and include the results in Figure \ref{fig:dna_lambda}.

\subsection{Comparison to PPO and PPG}

\begin{figure}[t]
    \centering
    $\vcenter{\hbox{
    \includegraphics[height=2.5in]{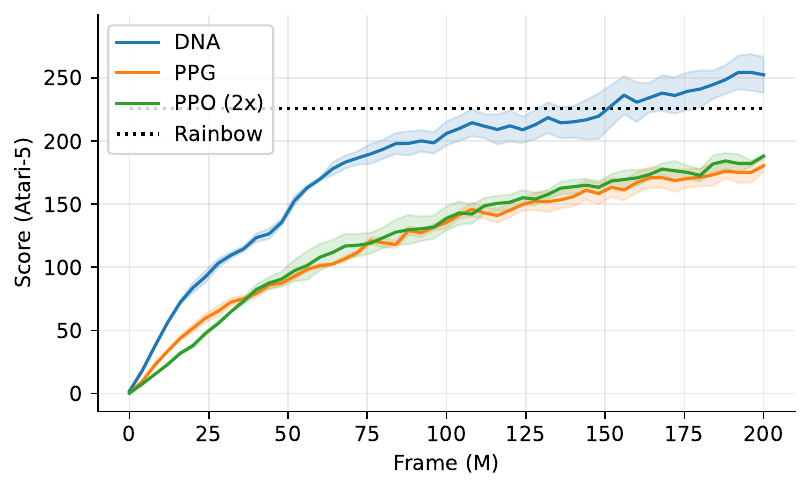}}}$
    $\vcenter{\hbox{
    \includegraphics[height=2.0in]{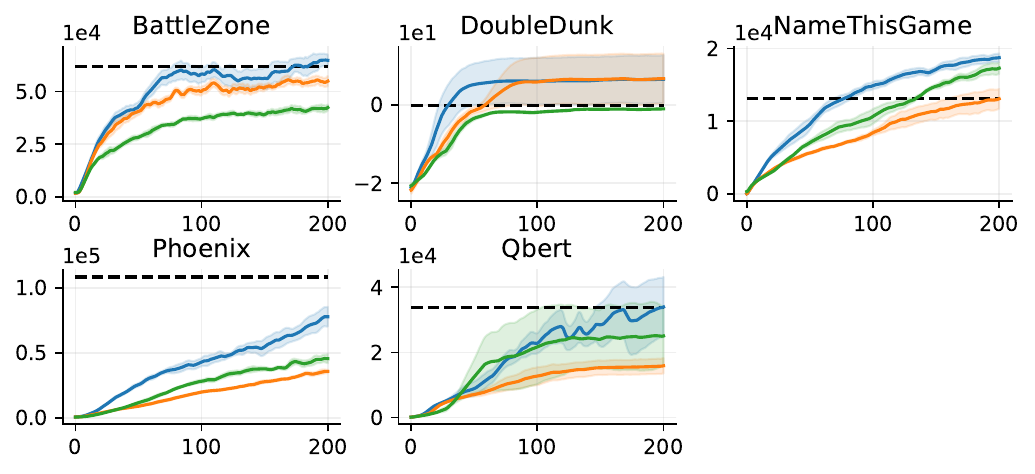}
    }}$
    \caption{Top: Results on the Atari-5 benchmark. DNA outperforms PPO by a wide margin, and even exceeds Rainbow DQN despite the more difficult settings used in our experiment. Bottom: Individual training curves for each of the five games.}
    \label{fig:final_plot}
\end{figure}

We found DNA to outperform both PPO and PPG on the Atari-5 dataset by a wide margin (Figure \ref{fig:final_plot}). We give Atari-5 scores, as well as training plots for each individual game. We also include references for Rainbow DQN, although we note that these scores were generated under the simpler deterministic settings.

\section{Discussion}
\label{sec:discussion}

\paragraph{Reduced Training as an Alternative to Experience Replay}

We found that \textit{less} training, both on policy and value updates, typically improves our agent's performance. One explanation for this could be, as \cite{cobbe2021phasic} have suggested, that in the case of value learning, this could be due to overfitting to recent experience.\footnote{This is plausible, especially with high variance returns.} In terms of policy updates, where the largest performance degradation was observed, this could be due to increased training epochs causing a large change in the policy, which may suggest a better constraint than PPO's clipping is required.\footnote{It is worth noting that PPO does clip the entropy bonus, and therefore updates with a non-zero entropy bonus could make arbitrarily large changes to the policy regardless of the choice of $\epsilon$.} That being said, limiting the training to one or two epochs provides straightforward solution to the problem while having the additional benefit of decreasing the computational resources required to train an agent. 

\paragraph{Choice of $\lambda$}

Our results give evidence for the benefit of selecting $\lambda_{\pi} < \lambda_V$. As expected, picking \textit{lower} values of $\lambda_\pi$ reduces the policy gradient noise. However, counterintuitively, noise on value learning reduces with \textit{higher} values of $\lambda_V$. That is, policy learning prefers low \textit{variance}, but value learning prefers low \textit{bias}. This result may explain why clipping, which introduces bias, performs poorly on value learning but quite well on policy learning.\footnote{By value clipping, we mean the value trust-region suggested by \cite{schulman2015high}, but which has been shown hurt performance by \cite{andrychowicz2020matters}.} Our results also suggest that an even higher choice of $\lambda_V$ than the $0.95$ we used in our experiments may be appropriate.



\paragraph{Update-to-Data Ratio}
\label{sec:update_ratio}

Our work also raises an interesting question about the update-to-data (UTD) ratio differences between value and policy learning. We found DNA to work best with only a single value network update but four policy network updates. This asymmetry could be due to the extremely high noise levels for policy updates. The single value update contrasts with other work where much higher value learning UTD ratios was found to be preferred \cite{aitchison2019optimal, chen2021randomized}.


\paragraph{Limitations}

Our paper's primary focus was improving PG results on discrete action vision-based problems, as this is where PG has traditionally unperformed Deep Q-learning. However, we do also provide an initial look into continuous control problems on the MuJoCo \cite{todorov2012mujoco} dataset in Appendix \ref{app:mujoco}.

\paragraph{Broader Impact}

In our work, we have shown that deep actor-critic reinforcement learning models can effectively solve complex vision-based, discrete-action environments, even without the use of large replay buffers. Surprisingly this can be achieved using \textit{less} not more policy and value network updates. Because Q-learning approaches generally produce deterministic policies, algorithms based on them may `lock in' to a decision based on minimal differences in outcome. This is not always the case with stochastic policies, which are able to randomly split over actions of roughly equal quality, which may lead to fairer outcomes. We do not foresee any direct negative societal impacts from this research. 

\paragraph{Future work} Our algorithm is already highly competitive with Rainbow DQN, and could likely be extended with orthogonal improvements such as adding recurrence \cite{hausknecht2015deep}, distributional value learning \cite{bellemare2017distributional}, improved feature encoders \cite{espeholt2018impala}, better exploration bonuses \cite{burda2018exploration}, and making use of a learned model \cite{schrittwieser2020mastering}. 

\section{Conclusion}


In this paper, we have highlighted noise level as a key difference between policy and value learning. We have introduced an algorithm that accounts for this order-of-magnitude difference by limiting \textit{negative} interference through the training of two independent networks for value and policy but retaining \textit{constructive} interference through a constrained distillation process. Furthermore, we have shown that the variance/bias tradeoff differs for value learning and policy gradient and that return estimation should cater for this. Together these changes result in a novel algorithm, DNA, that outperforms its predecessor PPO, and even surpasses the popular Q-learning approach Rainbow DQN, while under more challenging environmental settings.

\section*{Acknowledgments and Disclosure of Funding}
This research was supported by an Australian Government Research Training Program (RTP) Scholarship.

\newpage

\bibliography{references}



\newpage

\appendix

\section{Implementation Details}

The implementation details often matter with reinforcement learning \cite{engstrom2019implementation}. For this reason, full source code is provided in the supplementary material for all experiments. This section details some of the more important implementation decisions made, most of which match those found in \cite{andrychowicz2020matters}. We ran all experiments (PPO, DNA, PPG) using these same implementation choices to confirm that performance differences were not due differences in implementation.

\paragraph{Reward normalization} We normalized rewards such that returns have unit variance, as is common with PPO.\footnote{We also clipped normalized rewards to $[-5, 5]$ but found that this occurred exceedingly rarely, especially after the first 1 million frames.} Even though we used two separate models, and therefore had less reason to balance the magnitude of the value and policy loss, we still kept reward normalization so that distillation loss, and its interaction with the policy constraint, would be of a similar scale between environments.

\paragraph{Observation normalization} We also adopted observation normalization. Each state $s$ was normalized by $s' = \text{clip}((s - s_\mu) / s_\sigma, -3, 3)$ where $s_\mu$ was the element wide mean over states seen by the agent so far, and $s_\sigma$ was the standard deviation. Normalization constants were shared between the policy and value networks.

\paragraph{Repeat action penalty} We found that our policy would occasionally get stuck repeating a single action, causing the game to freeze until the time limit occurred. This could occur if the agent mistakenly thought it would get a slightly negative score for continuing and therefore acted to postpone that reward as long as possible.\footnote{An example of this would be the agent failing to press the reset button after losing a life in Breakout.} Q-learning algorithms, such as DQN, which make use of $\epsilon$-greedy, do not experience this problem so long as $\epsilon > 0$. To address this, we implemented a reward penalty of $0.25$ (normalized) if the agent repeated the same action more than 100 times. We leave finding a better solution to this problem for future work.\footnote{Adding some kind of exploration strategy, such as Random Network Distillation \cite{burda2018exploration} would likely solve this problem more elegantly.}

\paragraph{Integrating time and action information} We added a watermark to the least recent frame in the 4-frame stack indicating the proportion of time which has occurred as a `progress bar' as well as markers on each frame indicating which action the agent selected on the previous frame. Inclusion of time is necessary to avoid violation of the Markovian property in time-limited environments \cite{pardo2018time}. Action indicators were added to allow the agent to understand when it repeated the same action multiple times, which is not always possible to determine from the state itself (due to multiple actions causing identical outcomes).

\paragraph{Warmup / desyncing environments} When initializing our environments, we ran each of the parallel 128 environments for $t \sim U(1,1000)$ interactions with actions selected uniformly over the action space. This served two purposes: to provide initial normalization parameters and to desynchronize the environments. We found that if we did not do this, agents would terminate around the same time on some environments, causing parallel rollouts to become correlated. While we found this made very little difference to the agent's performance, it removed oscillating scoring artefacts found early in training in some environments.

\section{Hyperparameters and Environmental Settings}
\label{app:hyperparameters}

We selected initial hyperparameters from an initial coarse hyperparameter search on the Atari-3 validation set. In some cases, where only small differences in performances were observed, we prefered settings that had been used in previous papers or were likely to be more efficient. For example, our search found a mini-batch size of 256 optimal for value and distil updates. However, we selected 512 instead, as the difference was not large and found this a more computationally efficient mini-batch size when trained on a GPU. A full list of hyperparameters are given in Table \ref{tab:main_hps}. We also provide hyperparameters for our PPG experiments in Table \ref{tab:ppg_hps}. The environmental settings we used are given in Table \ref{tab:env_settings}. 

\begin{table}[h]
    \centering
    
    \begin{tabular}{l r r r }
    \toprule
        Setting & DNA & PPO & $\text{PPO}_\text{orig}$ \\
    \midrule
        Entropy bonus ($c_\text{eb}$)   & 0.01 & 0.01 & $\alpha \times 0.01$  \\ 
        Rollout horizon (N)             & 128 & 128 & 128 \\
        Parallel agents (A)             & 128 & 128 & 8 \\
        PPO epsilon $\epsilon$          & 0.2 & 0.2 & 0.1 \\
        Discount gamma ($\gamma$)       & 0.999 & 0.999 & 0.99 \\
        
        Learning Rate                   & $2.5 \times 10^{-4}$ & $2.5 \times 10^{-4}$ & $\alpha \times 2.5 \times 10^{-4}$ \\
        
        Policy lambda ($\lambda_\pi$)   & 0.95 & 0.95 & 0.95 \\ 
        Value lambda ($\lambda_V)$     & 0.95 & 0.95 & 0.95 \\ 
        
        Policy epochs ($E_\pi/E_\text{ppo}$)         & 2 & 2 & 3 \\
        Value epochs ($E_V$)            & 2 & - & - \\
        Distil epochs ($E_D$)           & 2 & - & - \\
        Distil beta ($\beta$)           & 1.0 & - & - \\
        
        Policy mini-batch size          & 2048 & 2048 & 256 \\ 
        Value mini-batch size           & 512 & - & - \\
        Distil mini-batch size          & 512 & - & - \\
        Repeated action penalty         & 0.25 & 0.25 & 0.25 \\ 
        Global gradient clipping        & 5.0 & 5.0 & 5.0 \\ 
        
    \bottomrule
    \end{tabular}
        
    \caption{Summary of hyperparameters found in coarse hyperparameter search. Epoch counts, and $\lambda_*$ values were further fine-tuned as detailed in the main study. For $\text{PPO}_\text{orig}$, $\alpha$ was linearly annealed over training from $[1, 0]$.}
    \label{tab:main_hps}
\end{table}

\begin{table}[h]
    \centering
    
    \begin{tabular}{l r r r }
    \toprule
        Setting & PPG & PPG (tuned) \\
    \midrule
        Policy epochs ($E_\pi$)           & 1 & 2 \\
        Value epochs ($E_V$)              & 1 & 1 \\
        Distil epochs ($E_D$)             & 0 & 0 \\
        Auxiliary epochs ($E_\text{aux}$) & 6 & 2 \\
        Auxiliary Period ($N_\pi$)        & 32 & 32 \\

    \bottomrule
    \end{tabular}
        
    \caption{Summary of hyperparameters used in the Phasic Policy Gradient experiments. All other hyperparameters were set according to the DNA settings in Table \ref{tab:main_hps}.}
    \label{tab:ppg_hps}
\end{table}

\begin{table}[h]
    \centering
    \begin{tabular}{l c c}
    \toprule
    Setting & Easy & Hard \\
    \midrule
    Terminal on Loss of Life    & True & False \\
    Action Space                & Minimal & Full \\
    Repeat Action Probability   & 0.0 & 0.25  \\
    \midrule
    Training frames             & \multicolumn{2}{c}{200M} \\
    Color / grayscale           & \multicolumn{2}{c}{Grayscale} \\
    Frame stacked               & \multicolumn{2}{c}{4} \\
    Action repetitions          & \multicolumn{2}{c}{4} \\
    Reward clipping             & \multicolumn{2}{c}{No} \\
    Episode timeout             & \multicolumn{2}{c}{108K} \\
    Resolution                  & \multicolumn{2}{c}{84 $\times$ 84} \\
    Noop Starts                 & \multicolumn{2}{c}{1-30} \\
    \bottomrule
    \end{tabular}
    \caption{Environmental Settings used in experiments. `Hard' mode settings follow best practice by \cite{machado2018revisiting}, `easy' mode correspond to those used in the Rainbow DQN paper \cite{hessel2018rainbow}, with the exception that we do not apply the domain specific reward clipping modification. Training  frames includes skipped frames, that is our agents performed 50M interactions with the environment.}
    \label{tab:env_settings}
\end{table}

\clearpage

\section{Ablation Study}
\label{app:ablation}

This appendix quantifies the performance contribution of several important components of DNA. We considered the following changes:

\begin{itemize}
    \item \textbf{DNA (mb=512)} Our analysis of noise levels suggested a much larger mini-batch size for policy updates than for value updates. We measure the impact of this change by evaluating DNA with mini-batch sizes for all three training objectives set to 512. 
    \item \textbf{DNA (no distil)}    
    Validation scores indicated an improvement in performance using distillation over no distillation. We verify that this result is replicated on our test set. 
    \item \textbf{DNA ($\lambda_V=\lambda_\pi=0.95$)} We measured the impact of using non-homogeneous values for $\lambda_V$ and $\lambda_\pi$ by testing with these both set to $0.95$. 
\end{itemize}

\begin{figure}[t]
    \centering
    \includegraphics[width=0.75\textwidth]{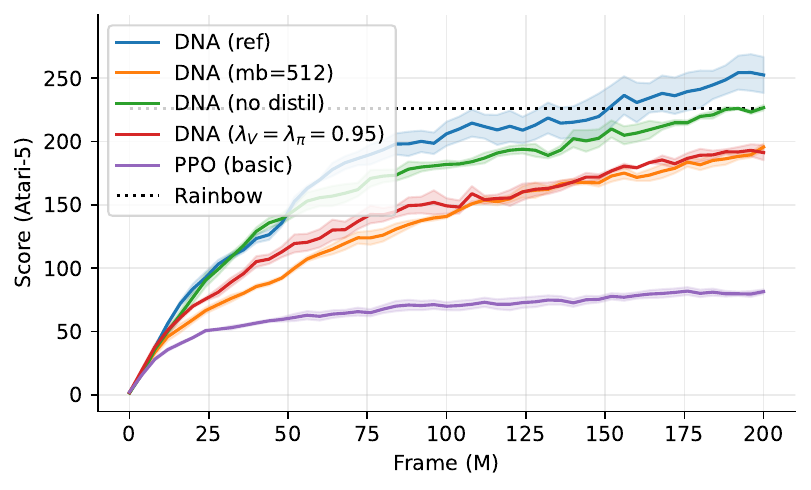}
    \caption{Training curves for the ablation studies. Shading indicates standard error over 3 seeds.}
    \label{fig:abl_results}
\end{figure}

We also include the reference run from the main study and a "PPO (basic)" run, which was a single network with the `Nature-CNN' encoder, and a mini-batch size of 512, and can be thought of as DNA with all novel components turned off. Results are provided in Figure \ref{fig:abl_results}, along with the score, and performance regression in Table \ref{tab:abl_results}. We found non-homogeneous values for $\lambda_V$ and $\lambda_\pi$, and a larger policy mini-batch size, to be the most significant changes, with distillation also providing some benefit. 

\begin{table}[h]
    \caption{Atari-5 scores for each of the ablation runs.}
    \label{tab:abl_results}
    \centering
    \begin{tabular}{l l l}
    \toprule
        Run & Atari-5 Score & Regression \\
    \midrule
         DNA (ref) & 252 & 0.0 \\
         DNA (no distil) & 226 & -10.2\% \\
         DNA (mb=512) & 195 & -22.5\% \\
         DNA ($\lambda_V=\lambda_\pi=0.95$) & 191 & -24.3\% \\
         PPO (basic) & 81 & -67.7\% \\
    \bottomrule
    \end{tabular}
    
\end{table}

\clearpage

\section{Noise Scale for Distillation Learning}
\label{app:distil_noise}

\begin{figure}[h]
    \centering
    \includegraphics[width=1.0\textwidth]{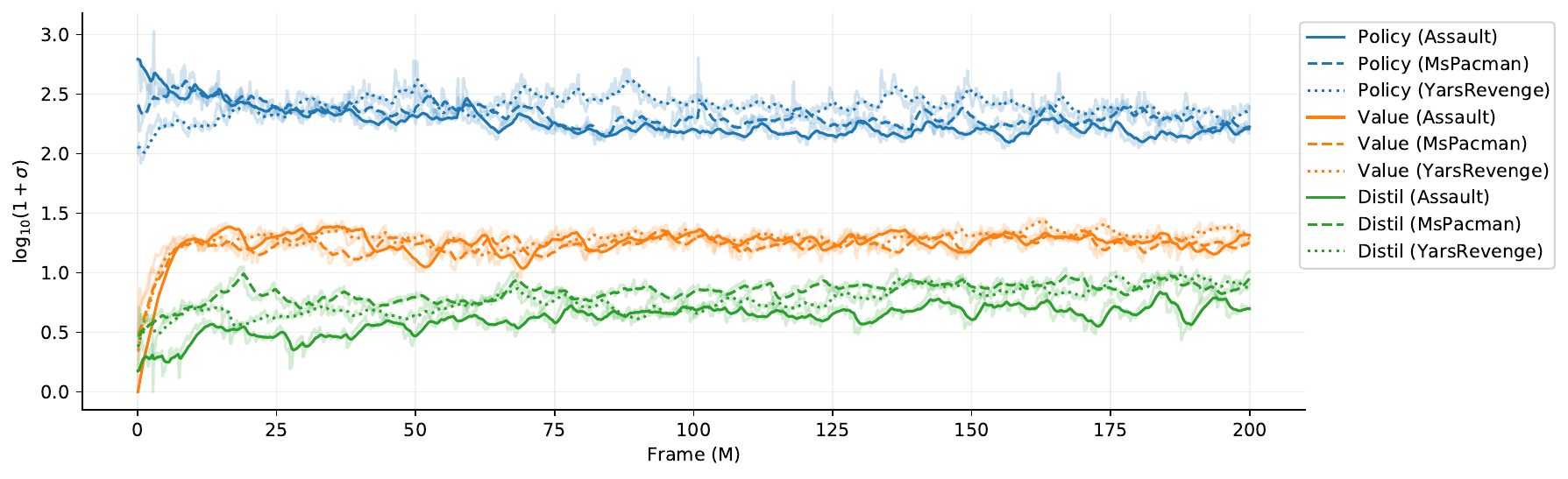}
    \caption{Noise level of the three tasks, policy learning, value learning, and distillation, over the three games in our validation set.}
    \label{fig:distil_noise}
\end{figure}

In this appendix, we present the noise scale results for distillation learning. We expected distillation to have a low noise level because the targets are drawn from the relatively noise-free value network estimations and not from the higher variance value targets. Our results in Figure \ref{fig:distil_noise} confirm this hypothesis. Of note is that the difference in noise scale between environments was much more significant for distillation loss than it was for the value or policy loss. These results indicate that distillation may benefit from smaller mini-batch sizes. However, the decreased efficiency of processing these smaller batches on GPU hardware may out-weigh any potential advantages.

\section{Proof of Relationship between GAE and TD($\lambda$)}
\label{app:td}

We provide a proof that calculating the General Advantage Estimate \cite{schulman2015high} is equivalent to calculating TD($\lambda$) returns, then subtracting the state value estimate. Concretely, for some $\lambda \in [0..1)$ and $\gamma \in [0..1]$ we have from \cite{schulman2015high}

\begin{align}
    \hat{A}_t^{(k)} := \sum^{k-1}_{l=0} \gamma^l\delta^V_{t+l} &= -V(s_t) + \Big( \sum_{i=0}^{k-1} \gamma^i r_{t+i} \Big) + \gamma^k V(s_{t+k})\\
    &= -V(s_t) + \text{NSTEP}^{(\gamma, k)}(s_t).
\end{align}

The GAE advantage estimate is defined as an exponentially weighted sum of these $A^{(k)}$'s as follows

\begin{align}
    \hat{A}_t^{\text{GAE}(\gamma, \lambda)} &:= \sum^{\infty}_{i=0} (1-\lambda)  \lambda^i \hat{A}_t^{(i+1)} \\ 
    &= (1-\lambda) \sum^{\infty}_{i=0} \lambda^i (-V(s_t) + \text{NSTEP}^{(i+1)}(s_t)) \\
    &= \Bigg[ (1-\lambda) \sum^{\infty}_{i=0} \lambda^i (-V(s_t)) \Bigg] +  \Bigg[ (1-\lambda) \sum^{\infty}_{i=0} \lambda^i \text{NSTEP}^{(i)}(s_t) \Bigg] \\
    &= -V(s_t) + \text{TD}^{(\gamma, \lambda)}(s_t)  
\end{align}

as required.

\section{Results under Rainbow DQN Style Environmental Settings.}
\label{app:easy}

In our main study, we compared DNA to PPO on the Atari-5 benchmark under the recommended settings given by \cite{machado2018revisiting}. Many prior results have been generated using the simpler, non-stochastic version of the environments and with domain-specific knowledge, such as loss of life as a terminal state and a custom clipped reward modifier. We evaluated DNA and our implementation of PPO here in the simplified environments and found a modest improvement under these settings. We provide these results for better comparison against previous works.

In these experiments we did not use reward clipping. Reward clipping is a domain-specific reward modification that reduces all positive rewards to +1 and all negative rewards to -1. We were concerned that clipping rewards would bias the algorithm, as the agent is optimizing a reward structure that may not match the true rewards of the game. It could be that reward clipping may be necessary for Deep Q-learning approaches to reduce high variance returns.\footnote{Or squashing the value function, see Appendix A of \cite{badia2020agent57}.} However, we have not found this to be an advantage over reward normalization for PPO or DNA. 

We found that DNA outperformed Rainbow DQN on all five environments, and obtained a better Atari-5 score after just 49M environmental frames. Proximal Policy Optimization, with a single policy update, also outperformed Rainbow DQN on this task.

\begin{figure}[H]
    \centering
    $\vcenter{\hbox{
    \includegraphics[height=2.5in]{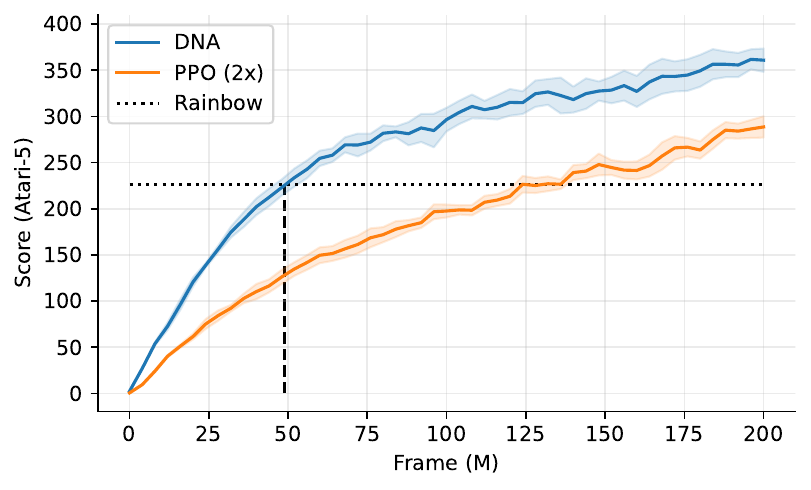}}}$
    $\vcenter{\hbox{
    \includegraphics[height=1.8in]{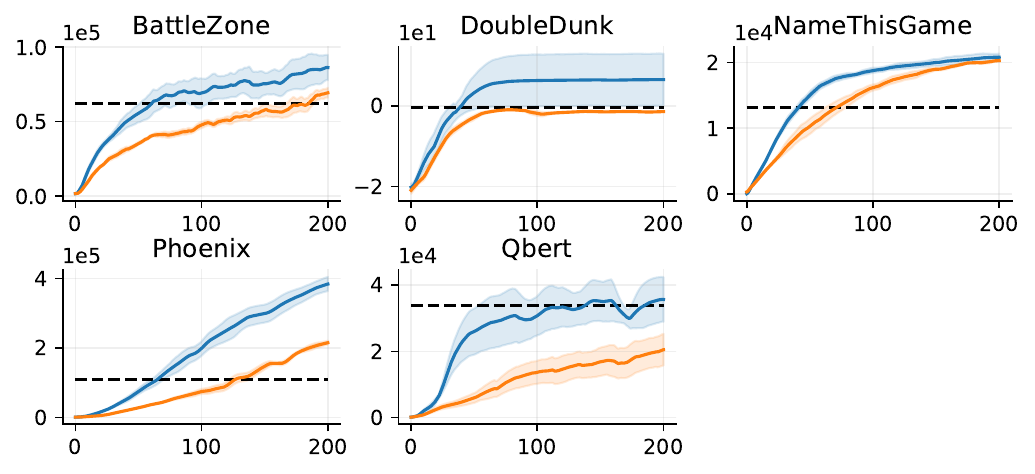}
    }}$
    \caption{Results on the Atari-5 benchmark, with `easy' environmental settings matched to those used by Rainbow DQN \cite{hessel2018rainbow}. Shaded regions indicate standard error over three seeds.}
    \label{fig:easy_plot}
\end{figure}

\section{Supplementary Results}

We investigated some supplementary questions. Specifically, we wanted to validate that the performance improvement of DNA compared to PPO and PPG did not result solely from the hyperparameter choices. We, therefore, evaluated PPO and PPG under a range of alternative hyperparameters on Atari-5 and note the results here.

\begin{figure}[h]
    \centering
    \includegraphics[width=0.49\textwidth]{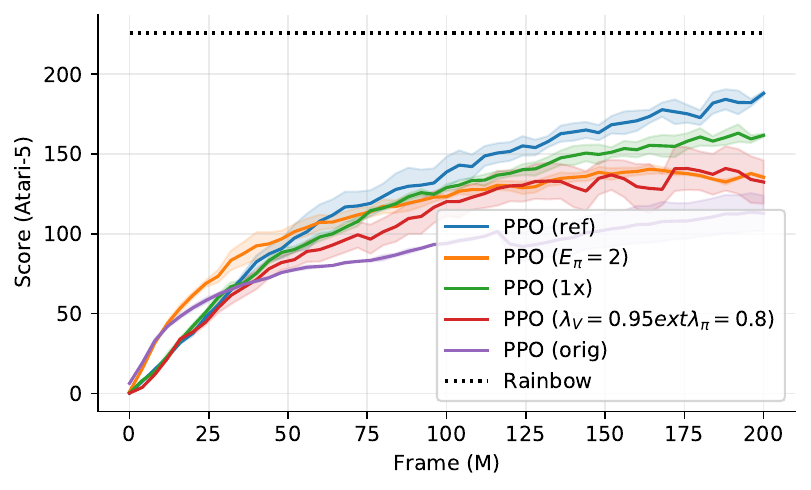}
    \includegraphics[width=0.49\textwidth]{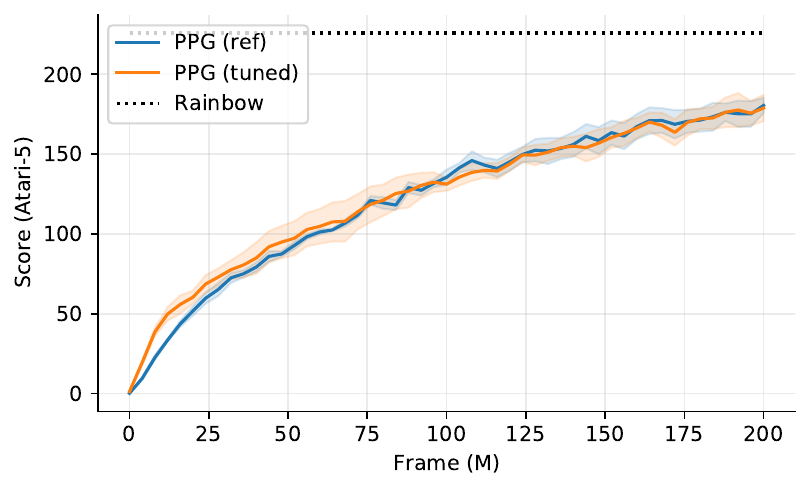}

    \caption{Training curves for the supplementary studies. Shading indicates standard error over 3 seeds. Reference runs are from the main study.}
    \label{fig:aux_results}
\end{figure}

We tested a variety of alternatives for PPO as described below. We found that none of the alternative settings resulted in improved performance (Figure \ref{fig:aux_results} \textit{left}).

\begin{itemize}

\item \textbf{PPO ($E_\pi=2$)}
Our tuning process found a single epoch optimal for PPO, but two epochs optimal for DNA, would PPO have performed better if it was given two epochs instead of one? We found that while initial performance was stronger, this change ultimately regressed the performance.

\item \textbf{PPO ($\lambda_V=0.95 \hspace{0.2em} \lambda_\pi=0.8$)} 
Using separate return estimations for advantages and value targets does not require a dual network setup. Therefore we checked if the performance of PPO can be improved by using the non-homogeneous $\lambda$ values used in our DNA experiment. We found these settings regressed performance.

\item \textbf{PPO (1x)}
In our experiments DNA and PPO used different network encoders (PPO used twice as many channels). It is possible that increasing the parameters made training more difficult for PPO? We tested PPO with the standard NatureCNN encoder, and found this change regressed performance.
     
\item \textbf{PPO (orig)}
Our settings for PPO deviated from those used by \cite{schulman2017proximal}. For completeness we also provide results using these settings. We found that while the these original settings performed well initially, they eventually unperformed the reference by a large margin. We also note that these settings took much longer to train (see Appendix \ref{app:time}).

\end{itemize}

We also evaluated PPG with alternative settings (Figure \ref{fig:aux_results} \textit{right}).

\begin{itemize}
\item \textbf{PPG (tuned)} In our main experiment we evaluated PPG using $E_\pi=1, E_V=1, E_\text{aux}=6$ taken from \cite{cobbe2021phasic}. These differ significantly from those used for DNA. We therefore reevaluated PPG using $E_\pi=2, E_V=1, E_\text{aux}=2$ which more closely match the settings used by DNA. We found this change had little impact on the performance of the algorithm.
\end{itemize}

\section{Training Time}
\label{app:time}

DNA makes use of two independent networks and three training phases, which may have a negative effect on training time. We examine this here. All times are approximate and for comparative purposes only. Rainbow DQN times are on different hardware and using a different codebase.

Our implementation of DNA ran very quickly and is faster than PPO when PPO is configured as per \cite{schulman2017proximal}. This is due to our use of more parallel agents (128 vs 8) coupled with a larger mini-batch size. We give approximate training times in Table \ref{tab:training_times} which were taken from a 24-core machine with four 2080-TIs. We found we could train 8 DNA models in 8-hours on our 4-GPU machine, giving a rate of 4 GPU hours per game learned. 

\begin{table}[h]
    \caption{Approximate training times for the algorithms used in this paper.}
    \label{tab:training_times}
    \centering
    \begin{tabular}{l r}
        \toprule
        Algorithm & GPU hours per game \\ 
        \midrule
        PPO (our settings)      & 3  \\
        DNA                     & 4  \\
        PPG                     & 4.5  \\
        PPO (\cite{schulman2017proximal} settings)      & 7.5 \\        
        Rainbow DQN & 240\footnote{There are faster ways of training DQN like algorithms, for example Ape-X \cite{horgan2018distributed}.}\\
        \bottomrule
    \end{tabular}
    
\end{table}

\section{Tuning for Proximal Policy Optimization}
\label{app:PPO_tuning}

We repeated the same hyperparameter sweep on PPO as we did for DNA for a fair comparison. We found that, like DNA, PPO also benefited greatly from reduced epochs during training. We present the results here for 1,2,3 and 4 epochs, along with a search over the choice for $\lambda$ used in the General Advantage Estimate. Results from the main study used the best performing model, which was found to be $\lambda=0.95$, and $\text{E}_\pi=1$. These hyperparameters differ from those used by \cite{schulman2017proximal}, and achieve a significant improvement in performance with less computation (see Appendix \ref{app:easy}, \ref{app:time}). 

We found the performance of PPO to plateau after a point in training which decreases with the number of training epochs. This is consistent with \cite{schulman2017proximal}, who trained for 40M frames and whose results show performance levelling off around 20M. However, when fewer epochs are used, performance continues to increase after these points. We also performed some quick experiments using partial epochs (0.5 and 0.75) but found these under-performed a single epoch and have not included the results here. We also found that the single network setup of PPO did not benefit as much as DNA from tuning the $\lambda$ parameter and that the commonly used $\lambda=0.95$ was optimal for our training set.

\begin{figure}[!h]
    \centering
    \includegraphics[width=0.49\textwidth]{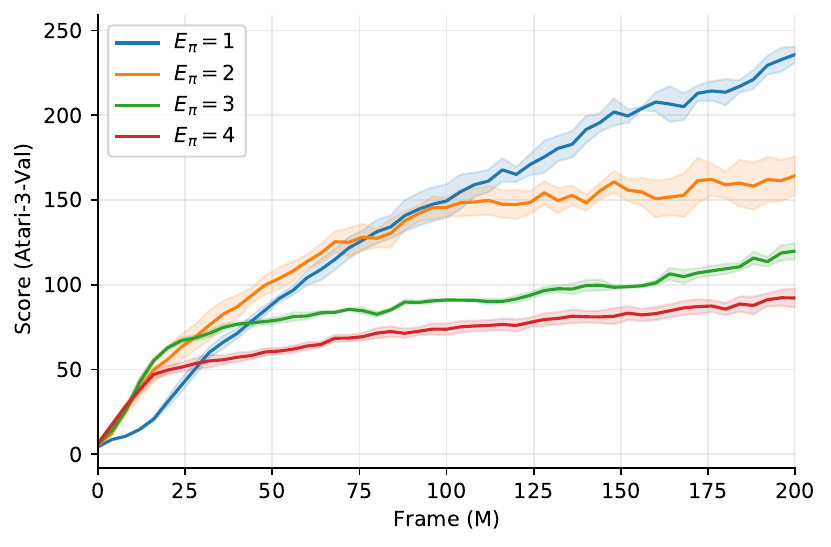}
    \includegraphics[width=0.49\textwidth]{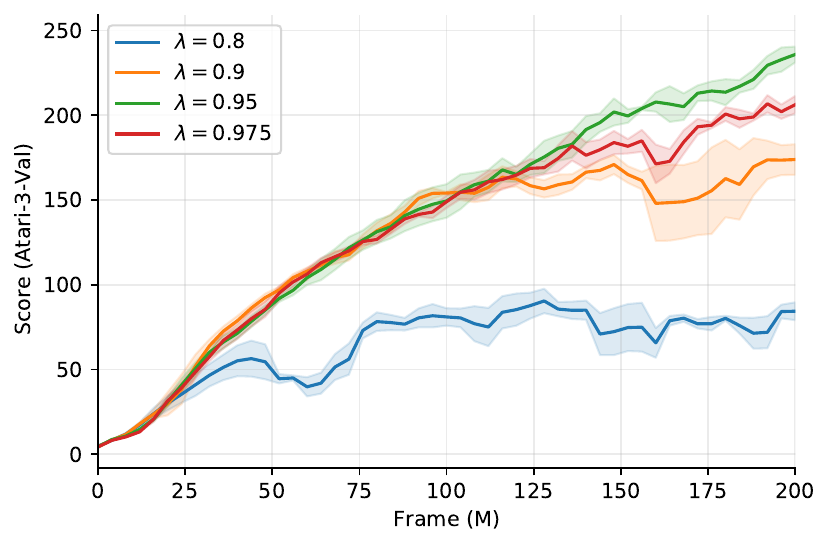}
    \caption{Training Curves for Proximal Policy Optimization over various epoch counts, and settings for $\lambda_\text{GAE}$. Shaded area indicates standard error over three seeds.}
    \label{fig:ppo_hps}
\end{figure}

\section{Distillation Targets}
\label{app:distil_targets}

\begin{figure}[h]
    \centering
    \includegraphics[width=0.45\textwidth]{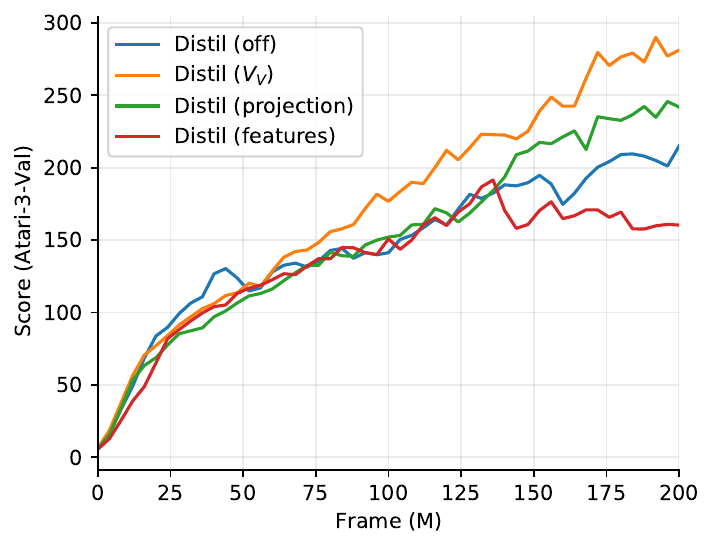}
    \caption{Performance of the four distillation strategies trailed. Only a single seed was used.}
    \label{fig:distil_target}
\end{figure}

We evaluated two distillation targets for our distillation phase: a \textit{random projection} into $\mathbb{R}^{16}$, and the value networks \textit{value estimates} as well as a third strategy, \textit{feature matching}.\footnote{That is, the distillation step minimized the mean squared error between the features outputted by the policy network and the features output by the value network. As with our other targets, gradients were only propagated through the policy network.} Random projection and value estimate outperformed the no distillation baseline, and feature matching underperformed the baseline. Results (Figure \ref{fig:distil_target}) are from a single seed on the Atari-3 validation dataset. In all cases, distillation was trained on two passes of trajectories sampled from the rollout and used a policy constraint.

\section{Additional Results on MuJoCo}
\label{app:mujoco}

We applied DNA to the robotics task MuJoCo \cite{todorov2012mujoco}. We used hyperparameters based on the work of \cite{schulman2017proximal}. However, we used the slightly different `v2' versions of the environments rather than the 'v1' versions used in their experiments. Scores are taken during training. PPO and DNA learned a per action standard deviation independent of the state. 

\begin{figure}
    \centering
    \includegraphics[width=0.9\textwidth]{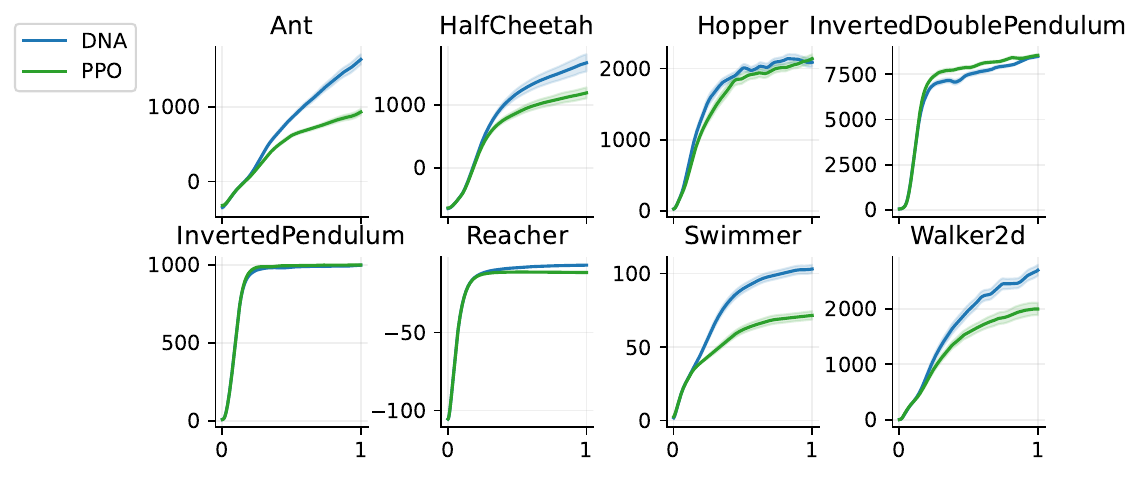}
    \caption{Results from the MuJoCo experiments for PPO and DNA over 30 seeds, with one standard error shown shaded.}
    \label{fig:mujoco}
\end{figure}

DNA shows an improvement over PPO on five of the eight of the environments tested (\textit{Ant}, \textit{HalfCheetah}, \textit{Reacher}, \textit{Swimmer}, and \textit{Walker2d}) (see Figure \ref{fig:mujoco}). In the remaining three environments, DNA and PPO produce similar results. We performed only basic hyperparameter tuning, using the \textit{Walker2D} environment. Hyperparameters for these experiments are given in Table \ref{tab:mujoco_hps}. 

Unlike in our Atari experiments, we found setting $\lambda_\pi$ to $0.8$ to produce poor results, and so reverted this setting back to $0.95$. We removed the KL divergence penalty for distillation and replaced it with the mean-squared error between the center of the Gaussian distribution outputs for the policy and value networks.

\begin{table}[h]
    \centering
    \caption{Hyperparameters used for MuJoCo. Hyperparameters follow closely to that of \cite{schulman2017proximal}. Agents where trained for one-million interactions. $\dag$ Learning rate was annealed linearly to $0.0$ over training.}
    \begin{tabular}{l r r}
    \toprule
        Setting & PPO & DNA \\
    \midrule
        Entropy bonus ($c_\text{eb}$)   & \multicolumn{2}{c}{0.01}\\ 
        Rollout horizon (N)             & \multicolumn{2}{c}{2048}\\ 
        Parallel agents (A)             & \multicolumn{2}{c}{1}\\ 
        PPO epsilon $\epsilon$          & \multicolumn{2}{c}{0.2}\\ 
        Discount gamma ($\gamma$)       & \multicolumn{2}{c}{0.99}\\ 
        Learning Rate                   & \multicolumn{2}{c}{$3.0 \times 10^{-4} \dag$}\\
        
        Policy lambda ($\lambda_\pi$)   & \multicolumn{2}{c}{0.95}\\ 
        Value lambda ($\lambda_V)$     & \multicolumn{2}{c}{0.95}\\  
        \midrule
        Policy epochs ($E_\pi/E_\text{ppo}$) & 10 & 10 \\
        Value epochs ($E_V$)            & - & 10 \\
        Distil epochs ($E_D$)       & - & 10 \\
        Distil beta ($\beta$)       & - & 1.0 \\
        
        Policy mini-batch size          & 64 & 64 \\ 
        Value mini-batch size           & - & 64 \\
        Distil/Aux mini-batch size      & -  & 64 \\
        Global gradient clipping        & 5.0 & 5.0 \\
    \bottomrule
    \end{tabular}
    \label{tab:mujoco_hps}
\end{table}

\clearpage

\section{Additional Results on Procgen}
\label{app:procgen}

We ran additional experiments on the Procgen benchmark \cite{cobbe2020leveraging}. Each environment within this benchmark requires learning across a set of 200 procedurally generated environments. Because of this, algorithms benefit from large replay buffers able to capture the diverse conditions under which the agent must act. Unlike PPG, DNA does not make use of replay buffer and so is not likely to perform well at this task. 

\begin{figure}[h]
    \centering
    \includegraphics[width=0.50\textwidth]{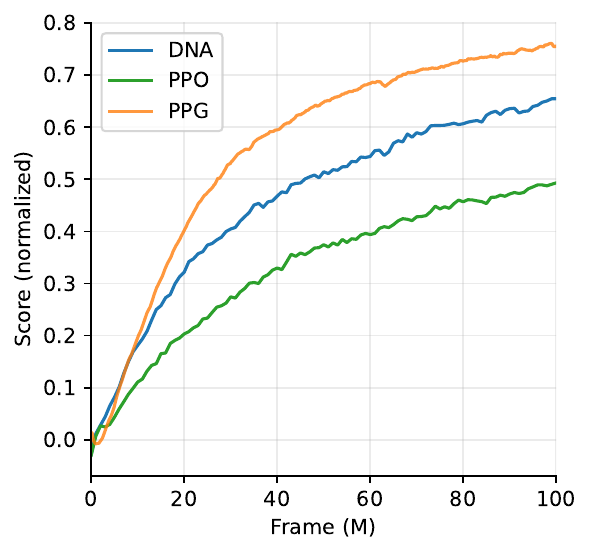}
    \caption{Results on the Procgen benchmark (hard distribution settings). Scores are average normalized score over all 16 environments. PPG results are taken from \cite{cobbe2021phasic}.}
    \label{fig:procgen}
\end{figure}

\begin{figure}[h]
    \centering
    \includegraphics[width=0.75\textwidth]{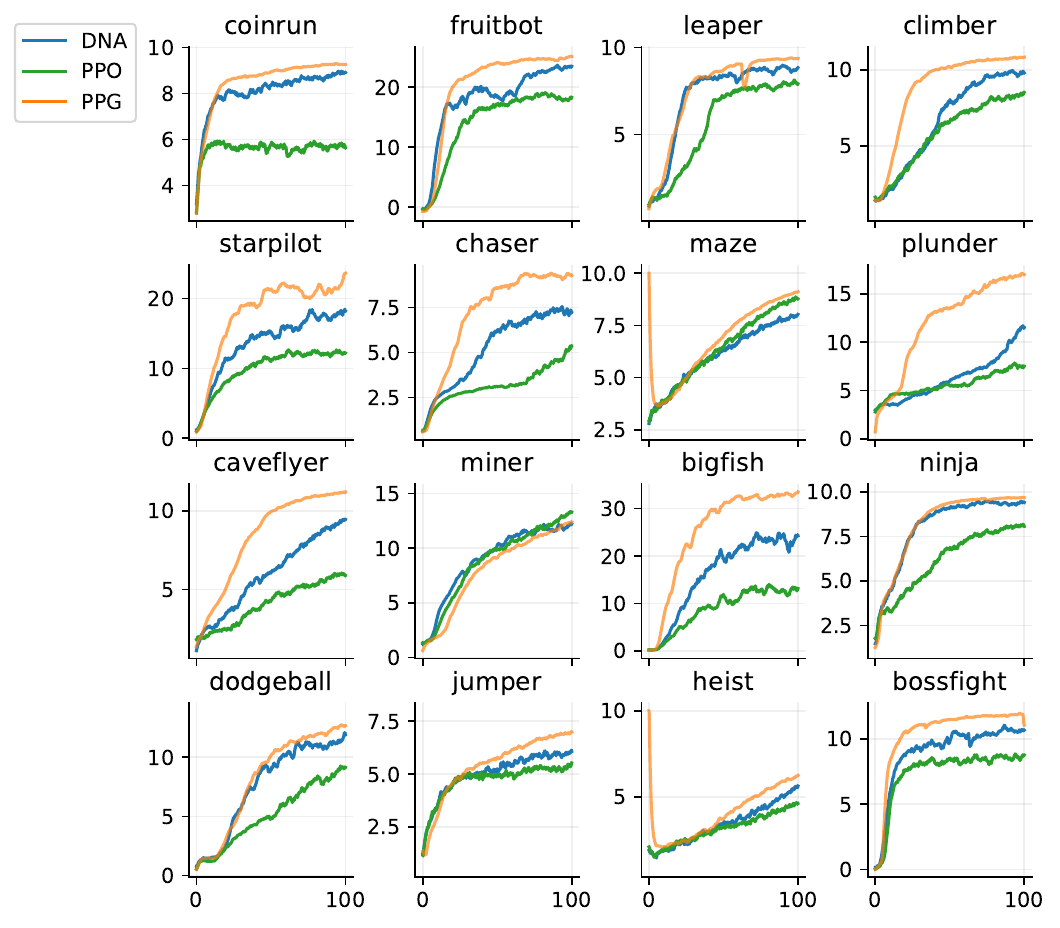}
    \caption{Results on each game in the Procgen benchmark (hard distribution settings). Scores smoothed for clarity. PPG results are taken from \cite{cobbe2021phasic} and are an average over three seeds. PPO and DNA are our results over one seed.}
    \label{fig:procgen_all}
\end{figure}

Despite this, we found DNA to be competitive with PPG on many of the environments tested (\textit{coinrun}, \textit{fruitbot}, \textit{leaper}, \textit{maze}, \textit{miner}, \textit{ninja}, \textit{dodgeball} and \textit{heist}), as shown in Figure \ref{fig:procgen_all}. When scores across all environments are normalized, DNA produces an average normalized score of $0.65$ compared to $0.49$ for PPO and $0.75$ for PPG (see Figure \ref{fig:procgen}). 

We kept hyperparameters close to that of \cite{cobbe2021phasic} and used the \textit{bigfish} environment to select one policy, one value, and two distil epochs as optimal for DNA. We note that these settings require the observations to be forwarded through a network only four times, compared to fourteen times in PPG. This makes DNA similar in computational requirements to PPO.\footnote{PPG's auxiliary phase requires forwarding through both the policy and value networks, whereas DNA's distillation update only requires a forward through the policy network.}

We generated results for DNA and PPO and used the results supplied by \cite{cobbe2021phasic} for PPG.\footnote{As recorded in the corresponding Github repository \url{https://github.com/openai/procgen}. We note some artefacts in the scores they supply, most notable in \textit{heist} and \textit{maze}. We believe this may be due to how their environments were initialized.} We also noticed a modest decline in the performance of our implementation of PPO compared to that of \cite{cobbe2021phasic} (0.49 vs 0.58). This could be due to our use of observation normalization. The difference in performance is especially apparent in the environment \textit{plunder} (for both DNA and PPO). This might indicate that performance could be improved further by adopting the fixed scaling preprocessing procedure used in PPG. We give hyperparameters for our experiments in Table \ref{tab:procgen_hps}.

\begin{table}[h]
    \centering
    \caption{Hyperparameters used for ProcGen. PPG hyperparameters are taken from \cite{cobbe2021phasic}.}
    \begin{tabular}{l r r r }
    \toprule
        Setting & PPO & DNA & PPG \\
    \midrule
        Entropy bonus ($c_\text{eb}$)   & \multicolumn{3}{c}{0.01}\\ 
        Rollout horizon (N)             & \multicolumn{3}{c}{256}\\ 
        Parallel agents (A)             & \multicolumn{3}{c}{256}\\ 
        PPO epsilon $\epsilon$          & \multicolumn{3}{c}{0.2}\\ 
        Discount gamma ($\gamma$)       & \multicolumn{3}{c}{0.999}\\ 
        Learning Rate                   & \multicolumn{3}{c}{$5.0 \times 10^{-4}$}\\
        
        Policy lambda ($\lambda_\pi$)   & \multicolumn{3}{c}{0.95}\\ 
        Value lambda ($\lambda_V)$     & \multicolumn{3}{c}{0.95}\\  
        Repeated action penalty         & \multicolumn{3}{c}{0.0}\\ 
        \midrule
        Policy epochs ($E_\pi/E_\text{ppo}$)         & 3 & 1 & 1 \\
        Value epochs ($E_V$)            & - & 1 & 1 \\
        Distil/Aux epochs ($E_D$)       & - & 2 & 6 \\
        Distil/Aux beta ($\beta$)       & - & 1.0 & 1.0 \\
        
        Policy mini-batch size          & 8192 & 8192 & 8192 \\ 
        Value mini-batch size           & - & 2048    & 8192 \\
        Distil/Aux mini-batch size      & -  & 512    & 4096 \\
        Global gradient clipping        & 5.0 & 5.0 & off \\  
    \bottomrule
    \end{tabular}
    \label{tab:procgen_hps}
\end{table}

\clearpage

\section{Additional Results on ALE}
\label{app:full}

Our main study used Atari-5 to allow enough time for seeded runs and because it provides an established training/test split. Because Atari-5 is a new benchmark, we thought it important to validate our algorithm's performance on the full 57-game suite. We, therefore, provide results for both PPO (2x) and DNA on all 57 games in the ALE using both the `easy' settings similar to \cite{hessel2018rainbow} and the more difficult settings used in our main study. 

We measured the median score as the median human-normalized score over the past 100-episodes and report the final median scores as the average for this measure over the last 10M frames (5\% of training frames). Individual game scores are also reported as the average over the final 10M frames.

\begin{figure}[h]
    \centering
    \includegraphics[width=0.49\textwidth]{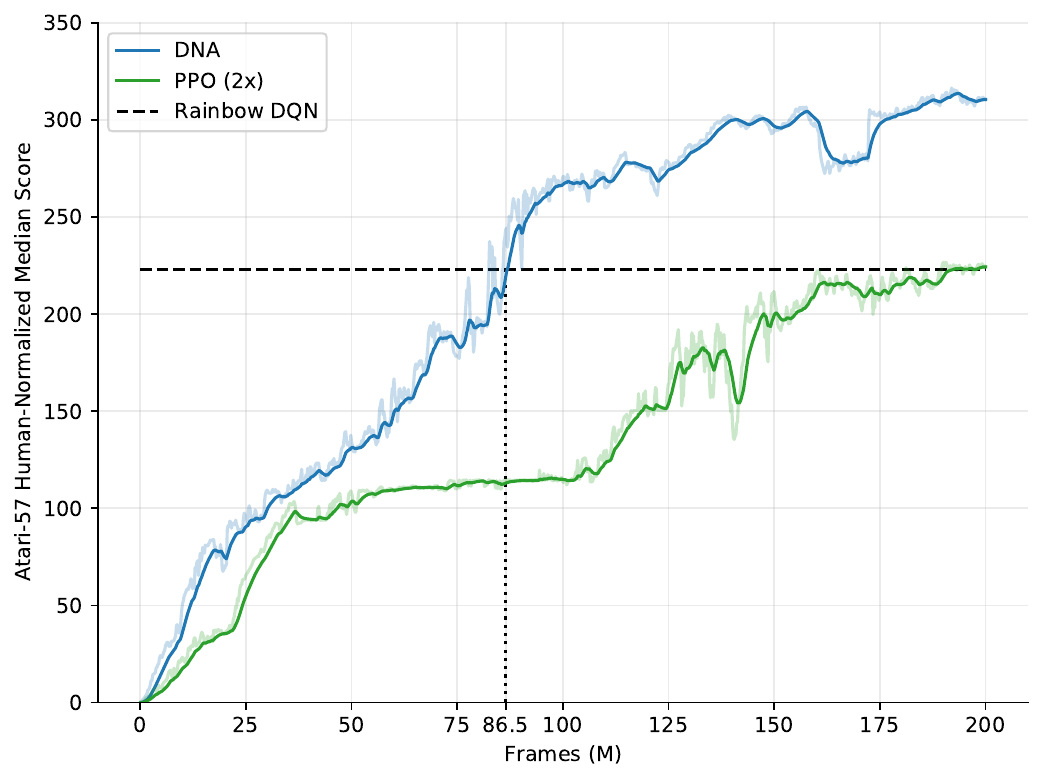}
    \includegraphics[width=0.49\textwidth]{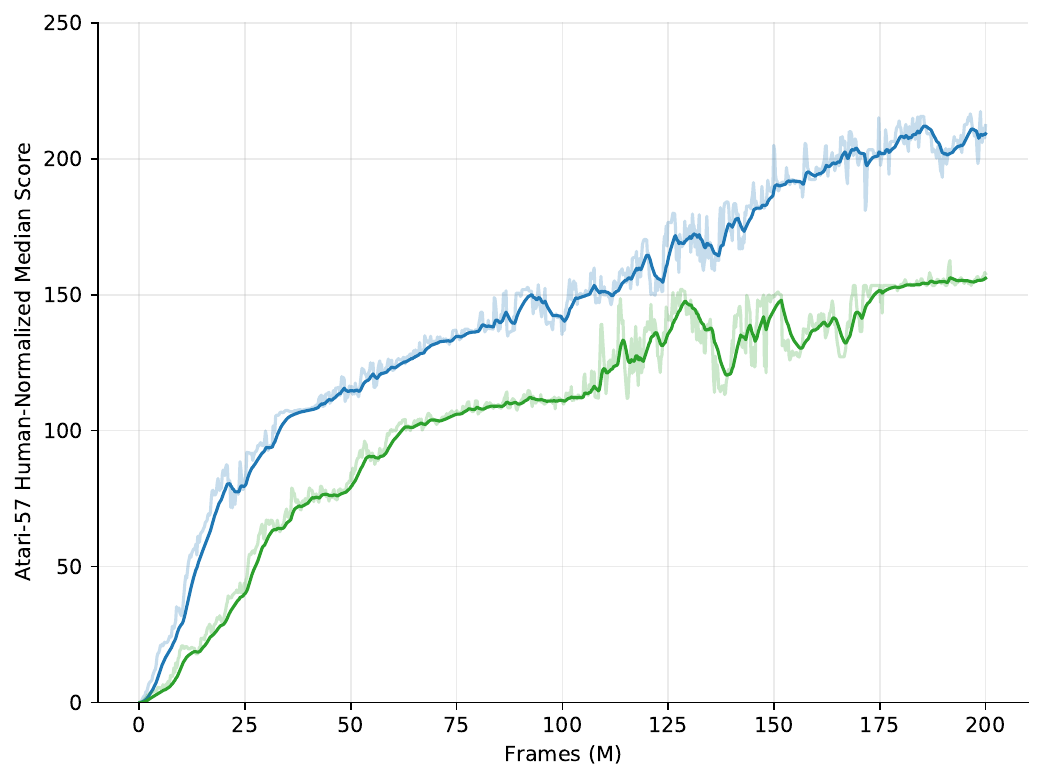}
    \caption{Median Score over all 57 games for DNA and PPO with 2x parameters. Left: performance under easy settings. DNA matches Rainbow DQN performance after just 86.5M frames. Right: performance under hard settings.}
    \label{fig:a57_median_plot}
\end{figure}

\begin{figure}[h]
    \centering
    \includegraphics[width=1.0\textwidth]{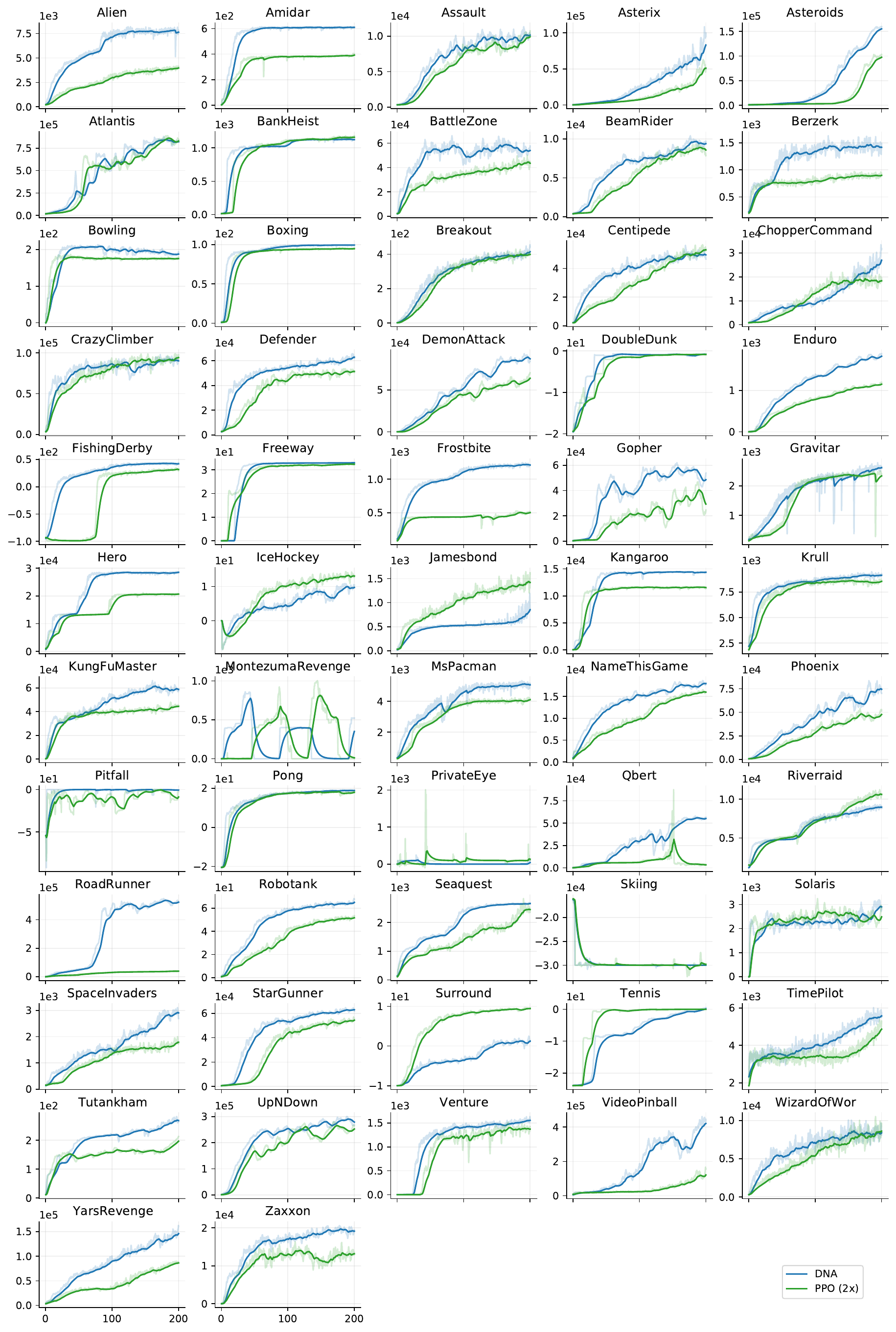}
    \caption{Training plots for DNA on all 57 games in the Atari-57 benchmark under `hard' settings. Results are from a single seed, with smoothed results in bold, and non-smoothed results shown faded.}
    \label{fig:a57_dna_hard}
\end{figure}

\begin{figure}
    \centering
    \includegraphics[width=1.0\textwidth]{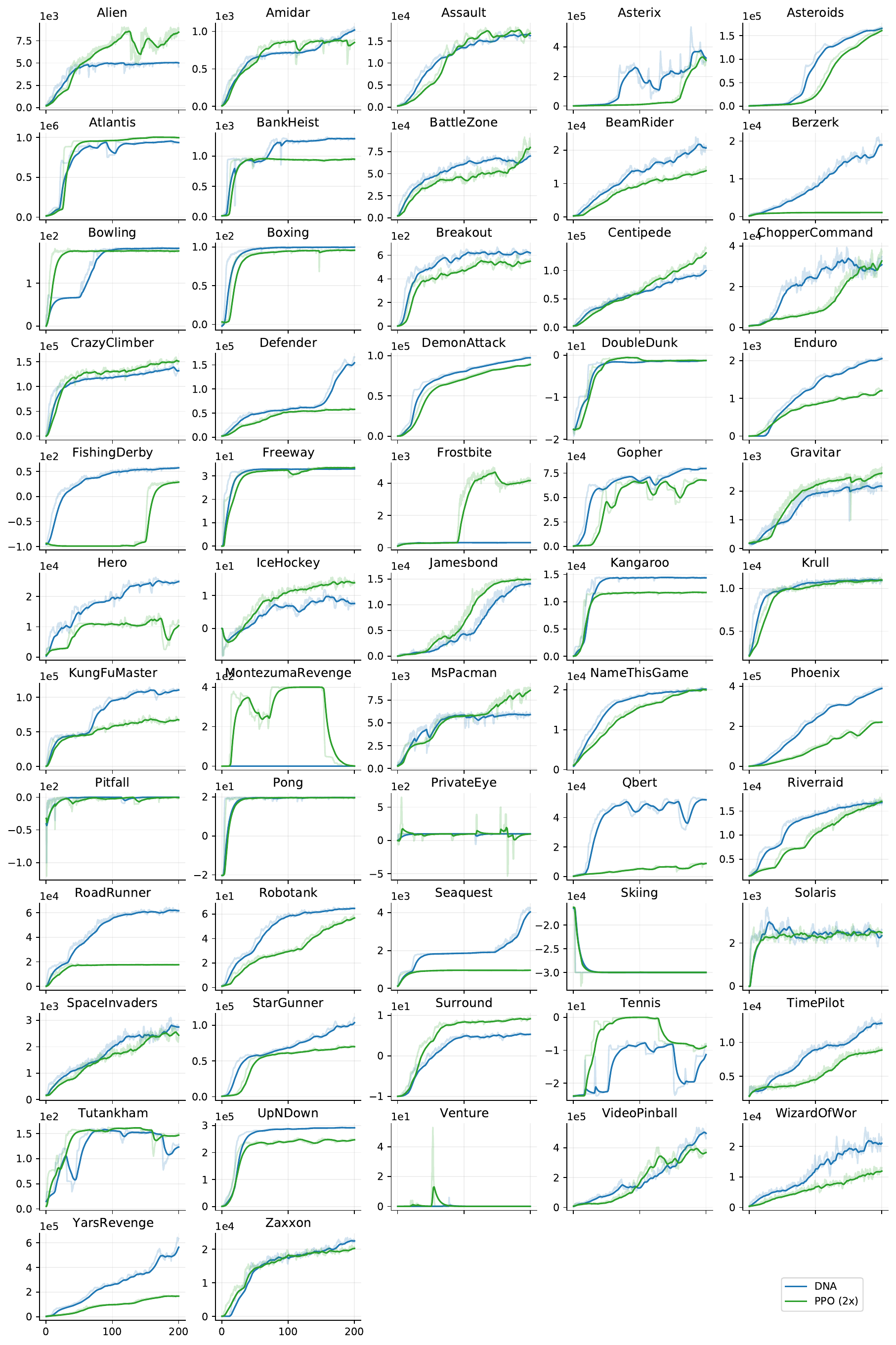}
    \caption{Training plots for DNA on all 57 games in the Atari-57 benchmark under `easy' settings. Results are from a single seed, with smoothed results in bold, and non-smoothed results shown faded.}
    \label{fig:a57_dna_easy}
\end{figure}

We found, under both the hard and the easy settings, DNA outperformed PPO by a wide margin (see Table \ref{tab:a57_summary}). DNA also outperformed Rainbow DQN on the easy settings after just 85.5M training frames (Figure \ref{fig:a57_median_plot}). Training plots are provided in Figures \ref{fig:a57_dna_hard}, \ref{fig:a57_dna_easy}. Results for each game are given in Tables \ref{tab:a57_full_hard}, \ref{tab:a57_full_easy}. Most surprising is that when trained with only a single policy epoch, a larger batch size, and more parallel agents, PPO becomes comparable to Rainbow DQN on the easy settings despite being a much simpler algorithm and being 80-times faster to train.

\begin{table}[h]
    \centering
    \caption{Summary of results on the Atari-57 benchmark.}
    \begin{tabular}{l r r r r}
    \toprule
         Algorithm &  Median (easy) & Median (hard) \\
    \midrule
         Rainbow & 223 & - \\
         PPO & 224 & 155 \\
         DNA (ours) & \textbf{311} & \textbf{207}\\
    \bottomrule
    \end{tabular}
    \label{tab:a57_summary}
\end{table}

\begin{table}[h]
    \footnotesize
    \caption{Final scores for all 57 games in ALE under `hard' settings.  Reported as mean score over the final 10M frames of training.}
    \label{tab:a57_full_hard}
    \centering
    \begin{tabular}{l r r r r}
\toprule
Game                 & Random     & Human      & PPO (2x)   & DNA         \\
\midrule
Alien                & 228        & 7,128      & 3,976      & \textbf{7,617} \\
Amidar               & 5.8        & \textbf{1,720} & 391        & 610        \\
Assault              & 222        & 742        & 10,098     & \textbf{10,282} \\
Asterix              & 210        & 8,503      & 52,958     & \textbf{85,070} \\
Asteroids            & 719        & 47,389     & 99,458     & \textbf{157,926} \\
Atlantis             & 12,850     & 29,028     & \textbf{816,033} & 813,564    \\
BankHeist            & 14.2       & 753        & \textbf{1,159} & 1,125      \\
BattleZone           & 2,360      & 37,188     & 44,227     & \textbf{54,462} \\
BeamRider            & 364        & \textbf{16,926} & 8,587      & 9,369      \\
Berzerk              & 124        & \textbf{2,630} & 897        & 1,429      \\
Bowling              & 23.1       & 161        & 175        & \textbf{187} \\
Boxing               & 0.1        & 12.1       & 94.7       & \textbf{99.4} \\
Breakout             & 1.7        & 30.5       & 399        & \textbf{416} \\
Centipede            & 2,091      & 12,017     & \textbf{53,676} & 49,444     \\
ChopperCommand       & 811        & 7,388      & 18,160     & \textbf{26,998} \\
CrazyClimber         & 10,780     & 35,829     & \textbf{94,695} & 89,864     \\
Defender             & 2,874      & 18,689     & 51,002     & \textbf{62,935} \\
DemonAttack          & 152        & 1,971      & 64,180     & \textbf{88,673} \\
DoubleDunk           & -18.6      & -16.4      & -0.8       & \textbf{-0.8} \\
Enduro               & 0.0        & 860        & 1,150      & \textbf{1,819} \\
FishingDerby         & -91.7      & -38.7      & 32.0       & \textbf{41.9} \\
Freeway              & 0.0        & 29.6       & 32.3       & \textbf{33.0} \\
Frostbite            & 65.2       & \textbf{4,335} & 497        & 1,211      \\
Gopher               & 258        & 2,412      & 28,785     & \textbf{46,348} \\
Gravitar             & 173        & \textbf{3,351} & 2,209      & 2,627      \\
Hero                 & 1,027      & \textbf{30,826} & 20,673     & 28,526     \\
IceHockey            & -11.2      & 0.9        & \textbf{12.9} & 9.4        \\
Jamesbond            & 29.0       & 303        & \textbf{1,444} & 861        \\
Kangaroo             & 52.0       & 3,035      & 11,556     & \textbf{14,367} \\
Krull                & 1,598      & 2,666      & 8,539      & \textbf{9,161} \\
KungFuMaster         & 258        & 22,736     & 44,648     & \textbf{58,895} \\
MontezumaRevenge     & 0.0        & \textbf{4,753} & 0.0        & 385        \\
MsPacman             & 307        & \textbf{6,952} & 4,102      & 5,067      \\
NameThisGame         & 2,292      & 8,049      & 16,050     & \textbf{18,155} \\
Phoenix              & 761        & 7,243      & 47,804     & \textbf{75,709} \\
Pitfall              & -229.4     & \textbf{6,464} & -10.4      & -0.7       \\
Pong                 & -20.7      & 14.6       & 17.9       & \textbf{18.9} \\
PrivateEye           & 24.9       & \textbf{69,571} & 129        & 36.5       \\
Qbert                & 164        & 13,455     & 3,273      & \textbf{54,706} \\
Riverraid            & 1,338      & \textbf{17,118} & 10,642     & 9,005      \\
RoadRunner           & 11.5       & 7,845      & 38,970     & \textbf{520,458} \\
Robotank             & 2.2        & 11.9       & 51.5       & \textbf{65.0} \\
Seaquest             & 68.4       & \textbf{42,055} & 2,494      & 2,655      \\
Skiing               & -17098.1   & \textbf{-4336.9} & -29553.4   & -29974.5   \\
Solaris              & 1,236      & \textbf{12,327} & 2,379      & 2,976      \\
SpaceInvaders        & 148        & 1,669      & 1,808      & \textbf{2,940} \\
StarGunner           & 664        & 10,250     & 54,488     & \textbf{62,760} \\
Surround             & -10.0      & 6.5        & \textbf{9.5} & 0.9        \\
Tennis               & -23.8      & -8.3       & 0.0        & \textbf{0.2} \\
TimePilot            & 3,568      & 5,229      & 4,907      & \textbf{5,554} \\
Tutankham            & 11.4       & 168        & 199        & \textbf{272} \\
UpNDown              & 533        & 11,693     & 250,253    & \textbf{280,014} \\
Venture              & 0.0        & 1,188      & 1,375      & \textbf{1,562} \\
VideoPinball         & 0.0        & 17,668     & 120,142    & \textbf{432,752} \\
WizardOfWor          & 564        & 4,756      & \textbf{8,834} & 8,480      \\
YarsRevenge          & 3,093      & 54,577     & 87,267     & \textbf{147,864} \\
Zaxxon               & 32.5       & 9,173      & 13,244     & \textbf{19,125} \\
\bottomrule
    \end{tabular}
\end{table}

\begin{table}[h]
    \footnotesize
    \caption{Final scores for all 57 games in ALE under `easy' settings.  Reported as mean score over the final 10M frames of training.}
    \label{tab:a57_full_easy}
    \centering
    \begin{tabular}{l r r r r r}
\toprule
Game                 & Random     & Human      & Rainbow DQN & PPO (2x)   & DNA \\        
\midrule
Alien                & 228        & 7,128      & \textbf{9,492} & 8,525      & 5,021      \\
Amidar               & 5.8        & 1,720      & \textbf{5,131} & 844        & 1,025      \\
Assault              & 222        & 742        & 14,198     & \textbf{16,688} & 16,293     \\
Asterix              & 210        & 8,503      & \textbf{428,200} & 321,207    & 323,965    \\
Asteroids            & 719        & 47,389     & 2,713      & 161,787    & \textbf{165,973} \\
Atlantis             & 12,850     & 29,028     & 826,660    & \textbf{997,292} & 932,559    \\
BankHeist            & 14.2       & 753        & \textbf{1,358} & 951        & 1,286      \\
BattleZone           & 2,360      & 37,188     & 62,010     & \textbf{82,834} & 71,003     \\
BeamRider            & 364        & 16,926     & 16,850     & 13,932     & \textbf{20,393} \\
Berzerk              & 124        & 2,630      & 2,546      & 1,083      & \textbf{19,789} \\
Bowling              & 23.1       & 161        & 30.0       & 175        & \textbf{181} \\
Boxing               & 0.1        & 12.1       & 99.6       & 95.6       & \textbf{99.9} \\
Breakout             & 1.7        & 30.5       & 418        & 553        & \textbf{626} \\
Centipede            & 2,091      & 12,017     & 8,167      & \textbf{131,062} & 100,194    \\
ChopperCommand       & 811        & 7,388      & 16,654     & \textbf{31,912} & 31,181     \\
CrazyClimber         & 10,780     & 35,829     & \textbf{168,788} & 151,937    & 131,623    \\
Defender             & 2,874      & 18,689     & 55,105     & 58,201     & \textbf{152,768} \\
DemonAttack          & 152        & 1,971      & \textbf{111,185} & 88,958     & 97,909     \\
DoubleDunk           & -18.6      & -16.4      & \textbf{-0.3} & -1.3       & -1.3       \\
Enduro               & 0.0        & 860        & \textbf{2,126} & 1,230      & 2,059      \\
FishingDerby         & -91.7      & -38.7      & 31.3       & 29.0       & \textbf{57.4} \\
Freeway              & 0.0        & 29.6       & \textbf{34.0} & 33.5       & 33.0       \\
Frostbite            & 65.2       & 4,335      & \textbf{9,590} & 4,190      & 320        \\
Gopher               & 258        & 2,412      & 70,355     & 67,850     & \textbf{80,104} \\
Gravitar             & 173        & \textbf{3,351} & 1,419      & 2,632      & 2,190      \\
Hero                 & 1,027      & 30,826     & \textbf{55,887} & 11,125     & 24,904     \\
IceHockey            & -11.2      & 0.9        & 1.1        & \textbf{13.8} & 7.2        \\
Jamesbond            & 29.0       & 303        & \textbf{19,480} & 14,947     & 14,102     \\
Kangaroo             & 52.0       & 3,035      & \textbf{14,638} & 11,687     & 14,373     \\
Krull                & 1,598      & 2,666      & 8,742      & \textbf{11,007} & 10,956     \\
KungFuMaster         & 258        & 22,736     & 52,181     & 67,657     & \textbf{110,962} \\
MontezumaRevenge     & 0.0        & \textbf{4,753} & 384        & 0.0        & 0.0        \\
MsPacman             & 307        & 6,952      & 5,380      & \textbf{8,712} & 5,894      \\
NameThisGame         & 2,292      & 8,049      & 13,136     & 20,053     & \textbf{20,226} \\
Phoenix              & 761        & 7,243      & 108,529    & 220,560    & \textbf{391,085} \\
Pitfall              & -229.4     & \textbf{6,464} & 0.0        & -0.5       & 0.0        \\
Pong                 & -20.7      & 14.6       & \textbf{20.9} & 19.6       & 19.7       \\
PrivateEye           & 24.9       & \textbf{69,571} & 4,234      & 99.9       & 100        \\
Qbert                & 164        & 13,455     & 33,817     & 8,836      & \textbf{52,398} \\
Riverraid            & 1,338      & 17,118     & \textbf{22,500} & 17,156     & 16,789     \\
RoadRunner           & 11.5       & 7,845      & \textbf{62,041} & 17,596     & 61,713     \\
Robotank             & 2.2        & 11.9       & 61.4       & 56.7       & \textbf{64.8} \\
Seaquest             & 68.4       & \textbf{42,055} & 15,899     & 957        & 4,146      \\
Skiing               & -17098.1   & \textbf{-4336.9} & -12957.8   & -29974.4   & -29974.0   \\
Solaris              & 1,236      & \textbf{12,327} & 3,560      & 2,513      & 2,225      \\
SpaceInvaders        & 148        & 1,669      & \textbf{18,789} & 2,497      & 2,731      \\
StarGunner           & 664        & 10,250     & \textbf{127,029} & 70,247     & 104,125    \\
Surround             & -10.0      & 6.5        & \textbf{9.7} & 9.1        & 5.3        \\
Tennis               & -23.8      & -8.3       & \textbf{0.0} & -8.8       & -10.9      \\
TimePilot            & 3,568      & 5,229      & \textbf{12,926} & 8,918      & 12,774     \\
Tutankham            & 11.4       & 168        & \textbf{241} & 147        & 127        \\
UpNDown              & 533        & 11,693     & 103,600    & 247,994    & \textbf{291,934} \\
Venture              & 0.0        & \textbf{1,188} & 5.5        & 0.0        & 0.0        \\
VideoPinball         & 0.0        & 17,668     & \textbf{533,936} & 359,099    & 505,392    \\
WizardOfWor          & 564        & 4,756      & 17,862     & 11,996     & \textbf{20,851} \\
YarsRevenge          & 3,093      & 54,577     & 102,557    & 166,670    & \textbf{564,513} \\
Zaxxon               & 32.5       & 9,173      & 22,210     & 20,330     & \textbf{22,588} \\
\bottomrule

    \end{tabular}

\end{table}

\section{Pseudocode for Generating Noise Scale Estimates.}
\label{app:noise_estimate}

In their paper, McCandlish et al. \cite{mccandlish2018empirical} define the noise scale as,
\begin{equation}
    \bn := \frac{\text{tr}(H\Sigma)}{G^THG}
\end{equation}
where at parameter values $\theta$, $G$ is the true gradient, $H$ is the true Hessian, and $\Sigma$ is the per-example covariance matrix. For large models, the calculation of the Hessian is not practical. We, therefore, use their simplified measure 

\begin{align}
    \bs := \frac{\text{tr}(\Sigma)}{|\vec{G}|^2}.
\end{align}

Even though this formulation makes the unrealistic assumption that the Hessian is a multiple of the identity matrix, empirical studies by McCandlish et al. have shown it to provide a surprisingly good approximation for $\bn$. 

Our method for generating estimates of $\bs$ follows closely that of \cite{mccandlish2018empirical} and is formalized in Algorithm \ref{alg:noise}. 
While the estimates for $\text{tr}(\Sigma)$, and ${|\vec{G}|^2}$ are both unbiased their ratio may not be. We mitigate this by averaging over multiple samples when calculating $\gbs$ and applying smoothing to our estimate for ${|\mathcal{G}|^2}$.


\begin{algorithm}[h]
    \caption{Estimate Noise Scale}
    \label{alg:noise}
    \begin{algorithmic}[1]
    \Function{EstimateNoiseScale}{
    
    $D$, a batch of data.
    
    $L$, a loss function.
    
    $\theta$, the model parameters.
    
    $N_\text{samples}$, the number of small mini-batches to use.
    
    $\bbb, \bbs$, the big and small mini-batch sizes.
    
    $|\mathcal{G}|^2_\text{old}$, the previous smoothed $|\mathcal{G}|^2$ value.
    
    $\alpha$, smoothing factor to use for $|\mathcal{G}|^2$
    }
    \State Define \text{SAMPLE}(x,y), to draw y samples from x without replacement.
    \State $D_\text{big} \gets \text{SAMPLE}(D, \bbb)$
    \State $|\gbl|^2 \gets |\nabla_\theta L_{D_\text{big}}(\theta)|^2$
    
    \State $|\gbs|^2 \gets 0$
    \For{$i = 1 .. N_\text{samples}$}
        \State $D_\text{small} \gets \text{SAMPLE}(D, \bbs)$
        \State $|\gbs|^2 \gets |\gbs|^2 + ({1}/{N_\text{samples}}) |\nabla_\theta L_{D_\text{small}}(\theta)|^2$
    \EndFor
    \State $|\mathcal{G}|^2_\text{new} \gets {\large(  \bbb |\gbl|^2 - \bbs |\gbs|^2   \large)}/{(\bbb - \bbs)}$
    \label{line:est}
    
    \State $\mathcal{S} \gets {\large( |\gbs|^2 - |\gbl|^2   \large)}/({1/\bbs - 1/ \bbb}) $
    
    \State $|\mathcal{G}|^2 \gets \alpha |\mathcal{G}|^2_\text{old} + (1-\alpha) |\mathcal{G}|^2_\text{new}$
    \Comment{Perform smoothing over $|\mathcal{G}|^2$, to reduce variance}
    \State $\bs \gets {\mathcal{S}}/{|\mathcal{G}|^2}$ 
    \State \Return \bs \Comment the (squared) noise scale estimate.
    \State \Return ${|\mathcal{G}|^2}$ \Comment{pass to next call of EstimateNoiseScale}
    
    \EndFunction
    \end{algorithmic}
\end{algorithm}

\end{document}